\documentclass{l4dc2026}

\usepackage{algorithm}
\usepackage{algpseudocode}

\usepackage{booktabs}
\usepackage{caption}

\usepackage{tikz}
\usetikzlibrary{tikzmark}
\usetikzlibrary{decorations.pathreplacing,calligraphy}
\usepackage{xcolor}

\usepackage[symbol]{footmisc}

\usepackage[most]{tcolorbox}
\usepackage{changepage}

\tcbset{
  cleanframebox/.style args={#1}{
    enhanced,
    arc=4pt,
    boxrule=0.8pt,
    colframe=#1,
    colback=white,
    frame style={dash pattern=on 2pt off 2pt},
    boxsep=0pt,
    left=0pt,
    right=2pt,
    top=4pt,
    bottom=4pt,
    before skip=2pt,
    after skip=2pt
  }
}

\title[Learning to accelerate distributed ADMM]{Learning to accelerate distributed ADMM \\using graph neural networks}
\usepackage{times}

\author{%
 \Name{Henri Doerks}\footnotemark[1]\Email{henri.doerks@math.uu.se} \\
 \addr Department of Mathematics, Uppsala University, Sweden
 \AND
 \Name{Paul H\"ausner}\footnotemark[1] \Email{paul.hausner@it.uu.se} \\
 \Name{Daniel {Hernández Escobar}} \Email{daniel.hernandez.escobar@it.uu.se} \\
 \Name{Jens Sj\"olund} \Email{jens.sjolund@it.uu.se} \\
 \addr{Department of Information Technology, Uppsala University, Sweden}
 \footnotetext[1]{Equal contributions.}
}

\begin{document}

\maketitle
\begin{abstract}
Distributed optimization is fundamental to large-scale machine learning and control applications.
Among existing methods, the alternating direction method of multipliers (ADMM) has gained popularity due to its strong convergence guarantees and suitability for decentralized computation.
However, ADMM can suffer from slow convergence and high sensitivity to hyperparameter choices.
In this work, we show that distributed ADMM iterations can be naturally expressed within the message-passing framework of graph neural networks (GNNs).
Building on this connection, we propose learning adaptive step sizes and communication weights through a GNN that predicts these hyperparameters based on the current iterates.
By unrolling ADMM for a fixed number of iterations, we train the network end-to-end to minimize the solution distance after these iterations for a given problem class, while preserving the algorithm’s convergence properties.
Numerical experiments demonstrate that our learned variant consistently improves convergence speed and solution quality compared to standard ADMM, both within the trained computational budget and beyond.\footnote[2]{The code is available at \url{https://github.com/paulhausner/learning-distributed-admm}.}
\end{abstract}

\setcounter{footnote}{0}
\renewcommand{\thefootnote}{\arabic{footnote}}

\section{Introduction}
The emergence of large-scale interconnected networked systems has driven the need for efficient distributed optimization algorithms across machine learning, control, and signal processing.
In these systems, each agent typically has access only to local information, yet must coordinate with neighbors to solve a global optimization problem.
The alternating direction method of multipliers (ADMM) is well-suited for this task and has gained increased popularity due to its flexibility and scalability \citep{boyd2011}.
A key feature of ADMM is that it can be implemented in a distributed fashion: agents perform computations locally and in parallel, then communicate updates to their
neighbors and incorporate the received information into their local variables.
While ADMM has a well-established convergence theory \citep{eckstein1990admm}, its convergence speed is highly sensitive to hyperparameter choices and can be slow in practice~\citep{zhang2019accelerating}.

A promising avenue for addressing these limitations lies in recent advances in machine learning.
In particular, the learning-to-optimize framework \citep{chen2022learning} aims to automate the design of faster algorithms by using data-driven techniques that enhance the convergence of iterative methods for a given problem class.
Building on this idea, we adopt an unrolling scheme \citep{monga2021algorithm}, where network parameters of our model are trained to minimize a loss function that reflects ADMM performance after a fixed number of iterations.
In distributed optimization problems, which consist of an interconnected network of agents, graph neural networks (GNNs) \citep{battaglia2018relational, scarselli2008graph} are a natural choice to parameterize the learning method as they can operate on graph-structured data that directly represents the underlying interconnected network.\looseness=-1

\begin{figure}[t!]
    \centering
\begin{tikzpicture}[scale=0.84,transform shape]
\tikzstyle{function} = [circle, draw, fill=white, text centered, rounded corners, font=\normalsize]
\tikzstyle{data}=[font=\large]


\begin{scope}
    \draw[color=red, line width=2pt] (0,0) rectangle (6, 4);
    \fill[gray!20] (0,0) rectangle (6, 4);
    \draw[white, line width=2pt] (3,0) -- (3,4);
    \node[function] (bottom1) at (1,1) {\texttt{msg}};
    \node[function] (top1) at (5,3) {\texttt{upd}};
    \node[data] (external11) at (7,3) {$ V'$};
    \node[data] (external12) at (7,1) {$ E$};
    \node[data] (V1) at (-1,3) {$ V^k$};
    \draw[->]  (V1) --  node[sloped, above, font=\tiny] {\hspace{0.5cm}{$(y_j^k,\lambda_j^k)$}} (bottom1);
    \draw[->] (V1) --  node[sloped, above,font=\tiny] {$(x_i^k,y_i^k,\lambda_i^k)$}(top1);
    \node[rectangle, draw, fill=white] at (3,2) (rectangle1) {\texttt{aggregate}};
    \node[data] (E1) at (-1,1) {${E}$};
    \draw[->] (rectangle1) -- node[sloped, above, font=\tiny] {${(\bar y_{\rightarrow i}^{k},\bar \lambda_{\rightarrow i}^{k})}\hspace{0.55cm}$} (top1);
    \draw[->] (E1) -- node[above, pos=0.25, font=\tiny]{$e_{ij}$} (bottom1);
    \draw[-] (bottom1) -- (rectangle1);
    \draw[->] (top1) -- node[sloped, above, font=\tiny] {\qquad \quad$x_i^{k+1}$}(external11);
\end{scope}

\begin{scope}[xshift=7.5cm]  
    \draw[color=blue,line width=2pt, dashed] (0,0) rectangle (6.3,4);
    \fill[gray!20] (0,0) rectangle (6.3,4);
    \draw[white, line width=2pt] (3,0) -- (3,4);
    \node[function] (bottom2) at (1,1) {\texttt{msg}};
    \node[function] (top2) at (5,3) {\texttt{upd}};
    \node[data] (external21) at (7.5,3) {$ V^{k+1}$};
    \node[data] (V2) at (-0.5,3) {$ V'$};
    \draw[->]  (V2) --  node[sloped, above, font=\tiny] {$\qquad x_j^{k+1}$} (bottom2);
    \draw[->] (V2) --  node[sloped, above,font=\tiny] {$(x_{i}^{k+1},\lambda_i^k)$}(top2);
    \node[rectangle, draw, fill=white] at (3,2) (rectangle2) {\texttt{aggregate}};
    \node[data] (E2) at (-0.5,1) {$ E$};
    \draw[->] (rectangle2) -- node[sloped, above, font=\tiny] {$\bar x_{\rightarrow i}^{k+1}\hspace{0.4cm}$} (top2);
    \draw[->] (E2) -- node[above, pos=0.75, font=\tiny]{$e_{ij}$} (bottom2);
    \draw[-] (bottom2) -- (rectangle2);
    \draw[->] (top2) -- node[sloped, above, font=\tiny] {\hspace{0.18cm}$(y_i^{k+1},\lambda_i^{k+1})$}(external21);
\end{scope}

\end{tikzpicture}
    \caption{Computation structure of iteration $k$ of the decentralized distributed ADMM algorithm as two message-passing steps.
    The initial node features ${V}^k$ contain the previous ADMM iterates $(x^k_i, y^k_i, \lambda^k_i)$ for each node $i$ and the non-zero edge features ${E}$ describe the weighted pairwise connections in the network which is fixed across iterations.
    After two message-passing steps -- each consisting of a message (\texttt{msg}), an aggregation (\texttt{agg}), and an update (\texttt{upd}) part -- the new iterates $(x^{k+1}_i, y^{k+1}_i, \lambda^{k+1}_i)$ of all nodes in the network are the output of the GNN.}
    \label{fig:ADMM_GNN}
    \vspace{-1em}
\end{figure}

In this paper, we show that classical decentralized and distributed ADMM iterations \citep{makh2017admm} can be implemented as a graph network using the message-passing framework \citep{battaglia2018relational} as we showcase in Figure~\ref{fig:ADMM_GNN} and further expand upon in Section~\ref{sec:gnn-admm}.
We explicitly exploit this connection and utilize GNNs to accelerate distributed ADMM.

In summary, our contributions in this paper are threefold:
First, we show the one-to-one correspondence between distributed ADMM iterations \citep{makh2017admm} and message-passing networks \citep{gilmer2017neural}, establishing an explicit 
algorithmic alignment \citep{velivckovic2021neural} between ADMM and GNNs.
Second, based on these insights, we develop and evaluate different approaches for learning instance-specific hyperparameters, namely the communication matrix and local or global step sizes, to accelerate the distributed ADMM algorithm.
These approaches correspond to graph-, node-, and edge-level learning tasks.
Finally, we propose a network architecture and a loss function to train the graph neural network in an end-to-end fashion by unrolling the ADMM iterations, ultimately resulting in an automated instance-specific hyperparameter tuning of ADMM.
In numerical experiments, we demonstrate that our learning-based approaches reduce the distance from the iterates to the minimizer compared to the baseline methods and can improve the convergence of ADMM even beyond the unrolling steps used during training.\looseness=-1

\section{Related work}
The goal of learning-to-optimize (L2O) approaches is to accelerate the convergence of parameterized optimization problems using machine learning techniques \citep{amos2023tutorial, chen2022learning}.
We focus on accelerating the ADMM algorithm for distributed optimization problems; for a detailed overview of the ADMM algorithm we refer to the survey by \citet{boyd2011}.

While L2O approaches can yield impressive results in practice \citep{andrychowicz2016learning}, they typically lack convergence guarantees and may fail to generalize, especially on unseen problem instances or when the underlying data distribution shifts \citep{amos2023tutorial}.
A popular approach to ensure convergence, which we adopt, is to learn the hyperparameters of an existing optimization algorithm.
Along those lines, \citet{sambharya2024learning} learn to warm start fixed-point algorithms, including ADMM, by unrolling the iterations and using automatic differentiation to compute the gradient through the iterations. 
In a follow-up work, \citet{sambharya2024hyper} learn hyperparameters of fixed-point iterations.
Similarly, \citet{NEURIPS2021_afdec700} use reinforcement learning to tune the hyperparameters of OSQP, an ADMM-based solver for convex quadratic programs.
These approaches have also been combined to solve quadratic programming problems with distributed ADMM \citep{saravanos2024deep}.
In contrast, we do not assume the presence of a central communication node. 
Further, we make no assumptions beyond the convexity of the objective function.
A similar unrolling technique for ADMM iterations was successfully applied by \citet{sjolund2022factorization} for nonnegative matrix factorization.
Another approach by \citet{noah2024distributed} learns the algorithm parameters directly through ADMM unrolling on a fixed set of agents.
We instead learn a network that predicts instance-specific hyperparameters of the optimization algorithm.
\citet{biagioni2020learning} use a recurrent neural network to directly predict the solution to ADMM which is used to warm-start the method.

Additionally, we propose a novel way to learn the communication matrix.
The predicted weighted Laplacian matrix can be seen as a form of learned preconditioning that stabilizes the iterates by improving problem conditioning  \citep{teixeira2015admm}.
Machine learning methods have previously been applied successfully to learn preconditioners for linear systems \citep{hausner2024neural, hausner2025learning}, imaging problems \citep{fahy2024greedy}, and gradient descent \citep{gao2025gradient}.\looseness=-1

GNNs have been used for other L2O purposes previously, as they can represent various optimization problems, such as quadratic programs \citep{chen2024expressive, schmidtobreick2025warm}, and can simulate the iterations of existing solvers such as interior point methods.
However, while GNNs can, in principle, approximate the solution mapping by simulating the solver iterations, they provide no convergence guarantees \citep{qian2024exploring}.
GNNs have also recently been applied to general distributed optimization.
In contrast to approaches that start with a learned algorithm and add constraints inspired by theory \citep{he2024mathematics}, \emph{we formulate distributed ADMM as a GNN operating on the communication network}.
Then, we parameterize the iterations with neural networks, allowing the GNN to learn on the problem class, thereby replacing extensive manual hyperparameter tuning with a data-driven procedure.
\looseness=-1

\section{Background}
\subsection{Distributed optimization}
\label{sec:distr_opt}
A distributed optimization problem consists of a network of $m$ interconnected agents, represented by the nodes $\mathcal{V} = \{ 1, 2, \dots, m\}$ of
a connected graph $\mathcal{G}=(\mathcal{V},\mathcal{E})$. An edge $(i,j)\in \mathcal{E}$ indicates that agent $i$ and $j$ can communicate. 
The agents jointly aim
to minimize the sum of local objective functions $f_i$ where each $f_i \colon \mathbb R^n\rightarrow \mathbb R \cup \{ + \infty \}$ is only known to agent $i$ and the data parameterizing the objective cannot be communicated.
Limited communication between agents can arise, for instance, due to limited networking capacity, as in edge networks \citep{dal2023q}, or due to privacy concerns \citep{chen2024privacy}.
In such scenarios, we must solve the problem in a distributed manner, which requires decoupling the objective minimization and adding a consensus constraint:\looseness=-1
\begin{align}
    \label{eq:distr_probl}
    \min_{x_1, x_2, \dots x_m} \quad  \sum_{i=1}^m f_i(x_i) \quad 
    \text{s.t.}  \quad x_1 = x_2 = \dots = x_m\,,
\end{align}
where each agent has its own local optimization variable $x_i\in \mathbb R^n$. The agents can reach consensus by communicating their local solutions to their neighbors as defined by the graph $\mathcal{G}$.

We assume that all $f_i$ are convex and that the objective \eqref{eq:distr_probl} is a proper, closed function. 
While there exist many different algorithms that solve this problem,
we focus on distributed ADMM for peer-to-peer networks, i.e., the agents communicate directly with each other.
In other words, we consider the \emph{decentralized} setting, which is more general than having a central hub or intermediary.\looseness=-1

\subsection{ADMM}
\label{sec:admm}
The alternating direction method of multipliers (ADMM) is an operator splitting technique that solves convex optimization problems of the form
\begin{equation}\label{eq:ADMM_problem}
    \min_{x,z}\;f(x)+g(z)
    \quad\text{s.t.} \quad Ax+Bz=c\,,
\end{equation}
where $x\in\mathbb R^p$, $z\in\mathbb R^q$, $A\in\mathbb R^{r\times p}$, $B\in\mathbb R^{r\times q}$, and $c\in\mathbb R^r$.
Originally proposed by \citet{Glowinski} and \citet{Gabay}, ADMM combines dual ascent, allowing parallel updates, and the method of multipliers, which converges under mild assumptions \citep{boyd2011}.\looseness=-1

\paragraph{Distributed ADMM}
Following the approach proposed by \citet{makh2017admm} we rewrite the distributed problem \eqref{eq:distr_probl} to match the general ADMM formulation \eqref{eq:ADMM_problem}.
In the rewritten form, the variable $x = [x_i]_{i=1}^m $ corresponds to the stacked individual variables of the agents and the function $f(x) = \sum_{i=1}^m f_i(x_i)$ is the global objective.
Further, the consensus constraint in \eqref{eq:distr_probl} is enforced by the equivalent formulation $(P \otimes I_n)x = 0$, where $P$ is a $m \times m$ communication matrix satisfying $\text{null}(P) = \text{span}\{1_m\}$.
This is satisfied by the (weighted) Laplacian $\mathcal{L}$ of a graph.
\looseness=-1

To allow parallel computation within ADMM, one can introduce an auxiliary variable $z$ and define $g$ as a convenient indicator function.
Moreover, by exploiting the inherent symmetry of the optimization variables, the edge-based auxiliary variables $z_{ij}$ can be replaced with corresponding node-based variables $y_i$, while the original dual variables are reparameterized as related variables $\lambda_i$.
This reparametrization reduces redundancy and leads to a more compact and efficient formulation.\looseness=-1

\paragraph{ADMM iterations}
Each iteration of the distributed ADMM algorithm updates all variables $x_i$, $y_i$, and $\lambda_i$ for each agent $i$.
The updates can be executed in parallel and require only information about the local objective and previous iterates of the agent's neighbors in the graph denoted by $\mathcal{N}(i) = \{ j : (i, j) \in \mathcal{E} \} \subset \mathcal{V}$.\footnote{
We do not allow for self-loops, so $i\notin \mathcal N(i)$. Further, since  $\mathcal G$ is undirected $(i,j)=(j,i)\in\mathcal E$.}
The local solution candidates $x_i$ is the solution to a regularized version of the local subproblem of agent $i$ and is computed as the minimizer of
\begin{equation}
    x_i^{k+1} =  \arg\min_{x_i} f_i(x_i)\,+ \hspace{-0.25cm}\underset{j\in \mathcal N(i)\cup\{i\}}{\sum}\hspace{-0.25cm}\left(( \lambda_{j}^{k})^T(P_{ji}x_i) +\frac{\alpha}{2}\|P_{ji}(x_i-x_i^k)+y_{j}^k\|_2^2\right)\,,
    \label{eq:xproblem}
\end{equation}
where $\alpha > 0$ is the global step size hyperparameter.
After communicating the updated iterates computed in \eqref{eq:xproblem} to all neighbors, the ADMM updates for
$y_i$ and
$\lambda_i$ are given by:
\begin{align}
    y^{k+1}_i = \frac{1}{d_i+1}\sum_{j\in \mathcal N(i)\cup\{i\}}P_{ij}x_j^{k+1}, \qquad \qquad 
    \lambda_i^{k+1} = \lambda_i^k+\alpha y_i^{k+1}.
    \label{eq:ylamproblem}
\end{align}
Note that the variable $y=[y_i]_{i=1}^m$ is a weighted residual for consensus since the constraint $(P\otimes I_n)x=0$ implies consensus on the components of $x_i$.
The full algorithm is given in Alg. \ref{alg:basic_admm}.

\paragraph{Convergence}
Under the assumptions on the objective function from Section \ref{sec:distr_opt}, the ergodic average of the computed local solutions $x_i$ converges for every global step size $\alpha > 0$ and any communication matrix $P$ with sublinear rate $\mathcal{O}(1/k)$ to the global solution.
However, both the global step size $\alpha$ and the communication matrix $P$, affect the convergence speed in practice, and optimal choices appear to be instance-dependent \citep{makh2017admm}.  The convergence results can be extended to the case where there exists not one but a sequence of step sizes \citep{eckstein1990admm}, or even node-level step sizes $\alpha_i>0$ \citep{barber2024node-level_alpha}. \looseness=-1
Moreover, the solution to \eqref{eq:xproblem} does not need to be exact to ensure convergence \citep{eckstein1990admm}.

\subsection{Graph neural networks}
Graph neural networks (GNNs) learn a feature representation for each node in a graph based on input features and the underlying graph topology.
In this paper, we focus on the framework of message-passing neural networks (MPNNs) which update the node features of a graph by aggregating information from the neighboring nodes and edges \citep{gilmer2017neural}.
Given edge weights $e_{ij}$ for $(i, j) \in \mathcal{E}$ represented by the weighted adjacency ${E}$ and initial node features $v_i^0$ for each node $i \in \mathcal{V}$, the latent node features $v_i^{l+1}$ are iteratively updated in each message-passing step $l$ by computing
\looseness=-1
\begin{equation}
    v^{l+1}_i = \texttt{update} \left( v^{l}_i, \texttt{aggregate}(\{ \texttt{message}(e_{ij} \cdot v^{l}_j : j \in \mathcal{N}(i)) \}) \right).
    \label{eq:mpnn}
\end{equation}
Here, the \texttt{update} and \texttt{message} functions can contain parameters that are learned during training, which are denoted by $\theta$.
We denote the node feature matrix in layer $l$ with ${V}^l$.
The \texttt{aggregate} function is typically permutation invariant, as there is no ordering of the graph neighbors, and can be implemented, for example, using the sum or mean function.
This makes the overall architecture permutation equivariant, so the ordering of the nodes does not affect the output of the network.

\section{Method}
Our key observation is that the distributed ADMM iterations introduced in Section~\ref{sec:admm} can be rewritten in the form of two message-passing steps of form \eqref{eq:mpnn}.
In Section~\ref{sec:gnn-admm}, we explicitly establish the connection between the ADMM iterations and GNNs.
Further, we describe how the hyperparameters of the iterations can be parameterized using neural networks.
In Section~\ref{sec:network}, we then describe how to train the parameters of this network using unrolled ADMM iterations.

\subsection{Learning ADMM via message-passing networks}
\label{sec:gnn-admm}
We start by deriving a one-to-one mapping from each distributed ADMM update step in \eqref{eq:xproblem} and \eqref{eq:ylamproblem} to a corresponding message-passing step.
Each step consists of a respective \texttt{message}, \texttt{aggregate}, and \texttt{update} function.
This correspondence is visualized in Figure~\ref{fig:ADMM_GNN}.

\paragraph{Input representation}
The distributed optimization problem can be directly encoded as the input to a GNN.
The underlying graph structure is naturally given by the communication network $\mathcal{G}$ from the problem formulation~\eqref{eq:distr_probl}.
The node features of the graph consist of the three ADMM iterates $(x_i,y_i,\lambda_i)$.
Furthermore, the non-negative edge weights $e_{ij}>0$ of the graph represent the weighted connectivity, that determine the strength of the connection between two nodes in the graph.
To obtain a valid communication matrix from the edge features, we construct the weighted Laplacian $P = \tilde D -  E$ where $\tilde D$ is a diagonal $m \times m$ matrix of the weighted degree for each node $\tilde D_{ii} = \sum_{j \in \mathcal{N}(i)} e_{ji}$ and $E$ is the symmetric matrix representation of the graph connectivity.

\paragraph{Message-passing steps}
We now reformulate the ADMM iteration in form of two message-passing steps.
Recall that in the MPNN framework in \eqref{eq:mpnn}, the \texttt{update} function only has access to the \texttt{aggregation} of the incoming messages and not the individual \texttt{messages}, as required in the original ADMM update step \eqref{eq:xproblem}. To define a valid \texttt{update} function, we rewrite \eqref{eq:xproblem} equivalently to 
\begin{align}\label{eq:node_update1}
     x_i^{k+1}=\underset{x_i}{\text{arg min}}\,f_i(x_i)+\bigl(P_{ii}\lambda_i^k+\bar \lambda_{\rightarrow i}^{\,k} +\alpha(P_{ii}y_i^k+\bar y_{\rightarrow i}^{\,k}) \bigl)^Tx_i +\frac{\alpha}{2}\hspace{-0.2cm}\sum_{j\in \mathcal N(i)\cup\{i\}}\hspace{-0.5cm}\|P_{ji}(x_i-x_i^k)\|_2^2.
\end{align}
In the rewritten form, $(\bar\lambda_{\rightarrow i}^k,\bar y_{\rightarrow i}^k)$ is the summation (\texttt{agg}) of all incoming \texttt{messages} to node $i$ that we define by $(P_{ji}\lambda_j^k, P_{ji}y_j^k)$.
Hence, in accordance with the MPNN framework, the refined \texttt{update} function in equation~ \eqref{eq:node_update1} now only depends on the aggregated features and the node's local information that is encoded in the node features. 
For details of the derivation, we refer to Appendix~\ref{app:alg}.
The updates for $y$ and $\lambda$ in equation~\eqref{eq:ylamproblem} can be carried out jointly within a second message-passing step, and their translation is straightforward.
It suffices to define the \texttt{message} between two adjacent nodes as $P_{ij}x_j^{k+1}$ and replace the respective part in \eqref{eq:ylamproblem} with the \texttt{aggregate} of these messages.\looseness=-1

Together, this yields a two-step message-passing network that we visualize in Figure \ref{fig:ADMM_GNN}.
Due to the equivalence to the original iteration in \eqref{eq:xproblem} and \eqref{eq:ylamproblem}, the network's output coincides with the updated iterates of a single distributed ADMM iteration.
\looseness=-1

\paragraph{Learned components}
The message-passing steps derived in the previous paragraph rigidly reproduce the decentralized distributed ADMM iterations, without any learnable parameters that could improve
performance.
In the following, we present several ways to introduce learnable parameters in the message-passing steps that still maintain convergence under assumptions stated in Section~\ref{sec:distr_opt}.
These approaches correspond to different levels of tasks for the GNN.\vspace{0.2cm}
\begin{adjustwidth}{1em}{0em}
\textbf{\textbullet\, Graph-level task}: We learn a \emph{global} step size by letting every node predict an $\alpha_i$ based on the local information and using the averaged $\alpha = \frac1m \sum_{i=1}^m \alpha_i$ for each node. In the distributed setting, this corresponds to introducing additional communication to reach consensus over the step size.\looseness=-1
\vspace{0.2cm} \\
\textbf{\textbullet\, Node-level task}:  Instead of choosing the same step size $\alpha$ for each node, we can learn an individual \emph{local} $\alpha_i$ for each node. This avoids additional communication between the nodes since the averaging step is omitted. However, each node is required to choose $\alpha_i$ without knowledge about the step sizes of the other nodes, making it conceptually more challenging.
\vspace{0.2cm} \\
\textbf{\textbullet\, Edge-level task}: For each edge in the network, we learn the positive \textit{edge weight} $e_{ij}$ between two connected agents. Based on the edge weights predicted before the first ADMM iteration, we use the corresponding weighted Laplacian $P$ as a fixed communication matrix for all steps.\looseness=-1
\end{adjustwidth}
\vspace{0.2cm}
For the first two tasks, the step size is predicted for each node individually using a neural network that receives as input the corresponding node features $(x_i, y_i, \lambda_i)$.
Additionally, we append the aggregated messages and the total number of nodes in the graph to the input.
No further information about the local objective $f_i$ is used as an input.
We apply instance normalization before passing the input through the network \citep{ulyanov2016instance}.
At the beginning of the algorithm, we can predict and change the step size in every iteration, resulting in a separate MLP for every change.
However, to ensure asymptotic convergence, we fix the step size after a predefined number of $L$ iterations.\looseness=-1\footnote{In practice, we often choose $L = K$.}

For the predicted edge weights, we use the local degree profile \citep{cai2018simple}, describing the local connectivity of the graph, as an input to an MLP.
The feature representations of the two nodes get stacked and passed through the network to predict a positive edge weight that preconditions the ADMM iterations.
To ensure invariance to the node order and symmetrize the edge weight graph, we sum the predicted edge weights for each directed edge.
To ensure the positivity of the learned algorithm hyperparameters $e_{ij}$ and $\alpha$, we use the \texttt{softplus} activation function.

\subsection{Network training}
\label{sec:network}
In real-world applications, distributed optimization problems are solved repeatedly for similar communication networks $\mathcal{G}$ and local objectives $f_i$, providing potential for problem-class specific acceleration.
We assume access to a dataset $\mathcal{D}$ consisting of problem instances that share an underlying structure to train our model to improve the average performance across the dataset.\looseness=-1

\paragraph{Unrolling}
We apply a technique called algorithm unrolling \citep{chen2022learning}, where the network is trained on a fixed number of iterations to maximize performance within this iteration budget.
In our case, we construct a GNN, consisting of $K$ distributed ADMM iterations via the message-passing scheme defined in the previous section.
This requires $2K$ message-passing steps for the GNN implementation.
Depending on the chosen learnable components, several MLPs are incorporated into the GNN to influence the \texttt{update} functions.
The parameters of all the neural networks are denoted by $\theta$.
The final output contains the approximate solution to the optimization problem~\eqref{eq:distr_probl} which now depends on these parameters.
The model is then trained using standard gradient-based learning methods that adjust the parameters $\theta$ to minimize a suitable loss function.

\paragraph{Loss function}
The loss function is designed such that it evaluates the GNN output $x^K=[x_i^K]_{i=1}^m$ only after the final $K$ unrolled iterations of ADMM.
Thus, we do not consider the learned ADMM performance after $k < K$ steps, nor how the iterates behave in subsequent iterations after $K$.

To train our model, we utilize a dataset $\mathcal{D}$, where each problem instance consists of a graph $\mathcal{G}$ and the per-node local objective functions $f_i$.
Additionally, we have access to the true minimizer $x^\star_d$ that solves problem \eqref{eq:distr_probl} for instance $d\in \mathcal D$ that can be computed offline efficiently since the problem is convex.
Further, we precompute $[\hat x^K_{d,i}]_{i=1}^m$, the approximate solution obtained by running the distributed ADMM iteration with default hyperparameters.
This is used to normalize the loss function in \eqref{eq:loss} and avoid over-weighting more involved instances in the training data.
We train the network to minimize the empirical risk on the training data by minimizing the loss function
\begin{equation}
    \ell(\theta; \mathcal{D})= \frac{1}{|\mathcal{D}|} \sum_{d=1}^{|\mathcal{D}|} {}\frac{1}{m_d} \left( \sum_{i=1}^{m_d}\frac{\|x_{d,i}^K(\theta)-x^\star_d\|_2^2}{\max
    \left( \|\hat x_{d,i}^K-x_d^\star\|_2^2, \varepsilon\right) } \, \right),
    \label{eq:loss}
\end{equation} 
using stochastic gradient-based optimization and automatic differentiation.
Here, $m_d$ is the number of agents in instance $d \in \mathcal{D}$, and $\varepsilon$ is a small positive constant that prevents division by zero. \looseness=-1

This loss function is regression-based as it relies on the true solution to the optimization problem~\eqref{eq:distr_probl} that can be computed apriori and is part of $\mathcal D$ \citep{amos2023tutorial}.
As such, the loss penalizes deviations from $x^\star_d$ without explicitly enforcing the consensus constraint or minimizing the global objective \eqref{eq:distr_probl}.
However, consensus is implicitly learned since all agents are trained to reach the same global solution.
The denominator does not depend on the network parameters but is used to normalize the loss function across problem instances.
Moreover, the normalized loss allows a natural interpretation of the network performance as it measures the network outcome relative to the baseline method with fixed hyperparameters.
Using an objective-based loss function instead of \eqref{eq:loss} requires adding an explicit term for consensus among the agents and a careful design of the normalization.\looseness=-1

\section{Experiments}
\label{sec:experiments}
\subsection{Problem setting}
We evaluate our method on two different use cases of distributed ADMM that are inspired by real-
world problems.
More details about the problem generation can be found in Appendix~\ref{app:problems}.\looseness=-1

\paragraph{Network average consensus problem}
The first problem we consider is network consensus where all agents have local information $b_i \in \mathbb{R}^n$ and the goal is to find the mean of all information across the network \citep{zhang2014lasso}.
Despite its simplicity, this problem is a classic benchmark that highlights important design choices in distributed optimization.
The objective function for each agent $i$ in the network is $f_i(x_i) = \| x_i - b_i \|^2$.

\paragraph{Distributed least-squares problem}
For the second use-case, we consider the distributed least-squares problem frequently arising in machine learning \citep{yang2020distributed}.
Each node has access to only $c$ samples of the data given by $B_i \in \mathbb{R}^{c \times n}$ and corresponding labels $b_i \in \mathbb{R}^{c}$.
The goal is to collaboratively find the least-squares solution with local objective:\footnote{In machine learning, the data is usually referred to as $(x, y)$ pairs and the parameters of the model as $\theta$. To be consistent with the ADMM framework above, we use $x\in\mathbb R^n$ as the regression parameters and $(B_i, b_i)$ as the data.} $f_i(x_i) = \| B_i x_i - b_i \|^2$.

\subsection{Results}
We train all our models for $K=10$ unrolling steps to minimize the loss function~\eqref{eq:loss} and compare the results with two baseline methods: (i) a version that uses the best performing $\alpha$ as a fixed global step size, and (ii) an adaptive heuristic that adjusts $\alpha_i$ dynamically. Additional details about the model training and baseline selection can be found in Appendix~\ref{app:training}. 
In Table~\ref{tab:resultsconsensus}, we show the results of different learned methods corresponding to individual learning components and a combined version in terms of their ability to minimize the loss function \eqref{eq:loss} and achieve consensus among the iterates.\looseness=-1

In the middle-left column, we show the key results obtained for $K=10$ iterations, which is the assumed computational budget used for the loss function in \eqref{eq:loss}.
We can see that all learned methods outperform the baselines, achieving a lower error across all learned methods in both experimental settings. This also holds for most methods in terms of consensus among the iterates, even though the loss only influences this indirectly. Overall, the combined method where we learn both edge weights and step sizes clearly performs best in both metrics. \looseness=-1

\paragraph{Additional iterations}
In the leftmost column of Table~\ref{tab:resultsconsensus}, we compare the results for $5$ steps.
Generally, the gap between the baseline methods and the learned methods is not as big as for the training objective.
However, already after $5$ iterations, the learned methods outperform the baselines in most cases, even though the training only evaluates the performance after $K=10$ steps.\looseness=-1

\begin{figure}
    \centering
    \includegraphics[width=\linewidth]{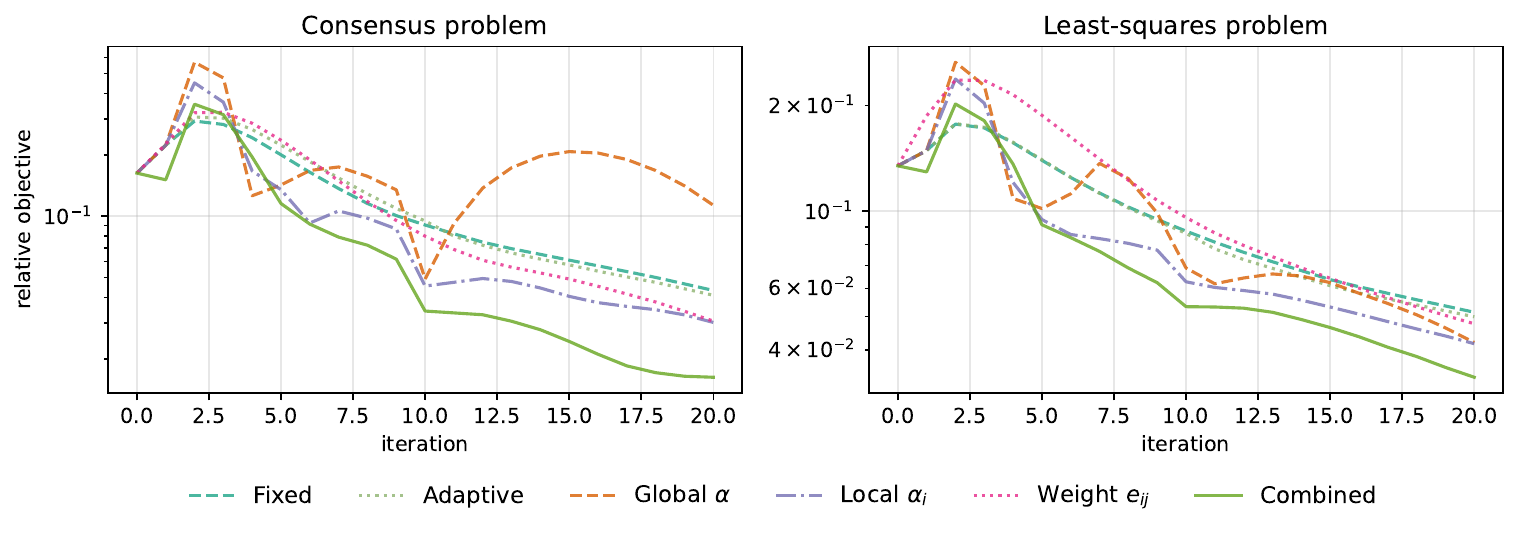}
    \vspace{-1em}
    \caption{Semi-log plots of the mean relative objective over all $100$ test instances for 20 iterations. The learned methods are only trained on the first $10$ iterations as in previous results.}
    \label{fig:function}
\end{figure}

In the middle-right column, we show the performance of the (learned) ADMM algorithm after $20$ iterations.
For the additional iterations, we fix the step size at $\alpha=1$ to guarantee convergence of ADMM and apply the same learned weighted Laplacian matrix.
Except for the global step size in the consensus problem, we observe improvements compared to the baselines.
Overall, the combined method demonstrates the strongest improvements.
Results for $100$ iterations, which is far beyond the training regime, are shown in the rightmost column. Apart from the combined method in the least-squares case, which appears to focus more on reaching consensus, all learned methods improve the baseline metrics.
In summary, although our method is trained to perform well only after the trained $10$ iterations, both earlier and later iterates still outperform the baseline methods in our experiments.\looseness=-1
\paragraph{Function value}
In addition, we evaluate the results in terms of the objective function value, measured by the relative objective at iteration $k$, i.e., the difference in objective values $\left| f(x^k) - f(x^\star) \right|$ normalized by the minimal value $|f(x^\star)|$.
In Figure~\ref{fig:function}, the relative objective is plotted for $20$ iterations.
We can clearly see that the performance benefit of the learned methods is largest after the $10$ unrolling steps used in training.
When additional iterations are executed, the model using a global $\alpha$ shows a large initial increase in the objective, likely due to overfitting on the number of unrolling iterations and freezing the step size. 
In the least-squares case, we also observe that the method using the learned communication matrix leads to a higher objective value for most of the iterations, even though the distance to the minimizer decreases.
However, when combining the communication matrix with the learned step size, the performance significantly improves, beyond simply adding the two individual contributions of these hyperparameters.
This shows that preconditioning still benefits the ADMM convergence, especially in synergy with learned step sizes.
More results comparing wall-clock times and out-of-distribution generalization of the methods can be found in Appendix~\ref{app:results}.\looseness=-1

\begin{table}[t]
    \centering
    \caption{Comparison of different methods of the ADMM algorithm after a fixed number of $k$ steps.
    Numbers reported are the average across the $100$ test instances.
    The error measures the average distance of the local solutions to the global minimizer $ \texttt{avg} \left( \| x_i^k - x^\star \|^2_2 \right) $ and the consensus gap measures the degree to which the consensus constraint is violated as $\texttt{avg} \left( \|x_i^k - \bar{x}^k\| \right) $, where $\bar{x}^k$ is the average solution at step $k$.
    We compare with a fixed baseline using the best constant step size and an adaptive method which applies a dynamic selection of node-level step sizes.
    In the combined version, we learn both individual step size parameters $\alpha_i$ and the edge weights simultaneously.
    Lower is better for all metrics. Best is \textbf{bold}, second best \underline{underlined}.
    }
    \label{tab:resultsconsensus}
    \resizebox{\textwidth}{!}{
    \begin{tabular}{lcccccccc}
    \toprule
    & \multicolumn{2}{c}{$k=5$} & \multicolumn{2}{c}{$k=10$ (\textbf{train})} & \multicolumn{2}{c}{$k=20$} & \multicolumn{2}{c}{$k=100$} \\
    \cmidrule(lr){2-3} \cmidrule(lr){4-5} \cmidrule(lr){6-7} \cmidrule(lr){8-9}
    \textbf{Consensus} & Error & Consensus & Error & Consensus & Error & Consensus & Error & Consensus \\
    \midrule
    Fixed & 18.54 & 11.02 & 8.99 & 6.48 & 2.26 & 2.02 & 0.144 & 0.144\\
    Adaptive & 19.02 & 12.45 & 8.86 & 6.98 & 2.13 & 1.94 & 0.140 & 0.139 \\
    Global $\alpha$ & 18.74 & 17.21 & 4.19 & 3.65 & 3.46 & 3.37 & 0.104 & 0.104 \\
    Local $\alpha_i$ & \underline{14.50} & 12.26 & \underline{3.05} & \underline{2.82} & \underline{1.17} & \underline{1.14} & 0.112 & 0.112 \\
    Weight $e_{ij}$ & 17.98 & \underline{10.96} & 7.39 & 5.47 & 1.52 & 1.40 & \underline{0.036} & \underline{0.036} \\
    Combined & \textbf{13.39} & \textbf{8.05} & \textbf{1.96} & \textbf{1.76} & \textbf{0.76} & \textbf{0.65} & \textbf{0.001} & \textbf{0.001} \\
    \midrule 
    \textbf{Least-squares} & Error & Consensus & Error & Consensus & Error & Consensus & Error & Consensus \\
    \midrule
    Fixed & 70.89 & 10.17 & 53.35 & 7.68 & 30.19 & 2.43 & 1.28 & 0.04\\
    Adaptive & 70.81 & 10.64 & 52.79 & 7.95 & 29.58 & 2.31 & 1.25 & 0.04 \\
    Global $\alpha$ & 61.32 & 23.41 & 27.12 & 9.21 & 14.93 & 3.48 & 0.67 & 0.04 \\
    Local $\alpha_i$ & \textbf{56.80} & 14.66 & \underline{23.79} & \underline{7.37} & \underline{12.96} & \underline{1.84} & \underline{0.62} & \underline{0.03} \\
    Weight $e_{ij}$ & 67.83 & 14.19 & 43.99 & 8.20 & 19.45 & 2.51 & \textbf{0.37} & \underline{0.03} \\
    Combined & \underline{58.97} & \textbf{8.83} & \textbf{18.24} & \textbf{5.42} & \textbf{11.79} & \textbf{0.95} & 1.94 & \textbf{0.01} \\
    \bottomrule
    \end{tabular}
    }
\end{table}

\vspace{-0.16cm}
\section{Discussion \& conclusion}
\vspace{-0.12cm}
We have shown how the distributed ADMM algorithm can be accelerated by expressing the updates in form of message-passing steps that can be parametrized to learn the step size and communication matrix.
Several other approaches could further exploit this connection.
In the following we highlight some of these potential extensions and discuss limitations of our approach.

\paragraph{Limitations}
ADMM has robust guarantees but is known to suffer from slow convergence when approaching the optimum. 
Since we intentionally stay within the ADMM framework, we cannot expect to overcome this inherent weakness. We only train the network to perform well within a fixed number of iterations $K$ across all instances.
This does not necessarily translate into improved performance on every individual instance, or faster global convergence beyond the $K$ iterations, which is a general limitation of unrolling \citep{arisaka2023principled}.
We observe this in our experiments, where the combined method leads to a larger error than the baselines when increasing the iterations significantly.
However, practical applications typically have a computational budget, making unrolling a useful technique for real-world problems.\looseness=-1
 \vspace{-0.08cm}

\paragraph{Future work}
We only augment ADMM by learning to predict the hyperparameters.
However, it is also possible to replace some of the ADMM steps completely.
For example, the update of the $x$-variable requires solving a potentially time-consuming optimization problem \citep{li2023learning}.
While it is tempting to replace this with the output of a neural network directly, ensuring convergence for such inexact methods is significantly more challenging \citep{banert2024accelerated}.
Additionally, the exchange of messages involving all iterates can be limited by the network bandwidth.
ADMM can be extended to only communicate compressed information without losing convergence \citep{chen2025greedy}.\looseness=-1

We focus on decentralized distributed ADMM \citep{makh2017admm} while many other formulations of distributed ADMM exist that can leverage graph neural networks.
In particular, distributed ADMM with a centralized communication hub can be mapped to message-passing with a global node, opening up many more synergies between distributed optimization and GNNs.\looseness=-1

\vspace{-0.04cm}
\paragraph{Conclusion}
In this paper, we demonstrate an equivalence between distributed ADMM and message-passing networks that, to our knowledge, has not been explicitly recognized before.
Based on this insight, we describe a learning-to-optimize approach that predicts instance-specific hyperparameters for distributed ADMM.
The model is trained by unrolling the algorithm and minimizing the normalized distance between the true minimizer and the network output for a specific problem distribution.
These learning tasks can be naturally expressed as different graph learning tasks leveraging the connection between ADMM and GNNs.
In numerical experiments, we validate that the learned algorithm outperforms the baseline methods and leads to faster convergence of ADMM.

\acks{The authors thank Daniel Arnström for helpful discussions and suggestions and the anonymous reviewers for their thoughtful comments. This research was funded in part by Sweden’s Innovation Agency (Vinnova), grant number 2022-03023, the Centre for Interdisciplinary Mathematics at Uppsala University (CIM), and supported by the Wallenberg AI, Autonomous Systems and Software Program (WASP) funded by the Knut and Alice Wallenberg
Foundation.
}

\bibliography{refs}

@inproceedings{schmidtobreick2025warm,
  title={Warm-starting active-set solvers using graph neural networks},
  author={Schmidtobreick, Ella J and Arnstr{\"o}m, Daniel and H{\"a}usner, Paul and Sj{\"o}lund, Jens},
  booktitle={8th Annual Conference on Learning for Dynamics and Control},
  journal={arXiv preprint arXiv:2511.13174},
  year={2026}
}

@article{biagioni2020learning,
  title={Learning-accelerated ADMM for distributed DC optimal power flow},
  author={Biagioni, David and Graf, Peter and Zhang, Xiangyu and Zamzam, Ahmed S and Baker, Kyri and King, Jennifer},
  journal={IEEE Control Systems Letters},
  volume={6},
  pages={1--6},
  year={2020},
  publisher={IEEE}
}

@inproceedings{dal2023q,
  title={Q-SHED: Distributed Optimization at the Edge via Hessian Eigenvectors Quantization},
  author={Dal Fabbro, Nicol{\`o} and Rossi, Michele and Schenato, Luca and Dey, Subhrakanti},
  booktitle={ICC 2023-IEEE International Conference on Communications},
  pages={4403--4408},
  year={2023},
  organization={IEEE}
}

@article{chen2024privacy,
  title={Privacy-preserving distributed optimization and learning},
  author={Chen, Ziqin and Wang, Yongqiang},
  journal={arXiv preprint arXiv:2403.00157},
  year={2024}
}

@software{jax2018github,
  author = {James Bradbury and Roy Frostig and Peter Hawkins and Matthew James Johnson and Chris Leary and Dougal Maclaurin and George Necula and Adam Paszke and Jake Vander{P}las and Skye Wanderman-{M}ilne and Qiao Zhang},
  title = {{JAX}: composable transformations of {P}ython+{N}um{P}y programs},
  url = {http://github.com/jax-ml/jax},
  version = {0.3.13},
  year = {2018},
}

@article{sambharya2024hyper,
  title={Learning algorithm hyperparameters for fast parametric convex optimization},
  author={Sambharya, Rajiv and Stellato, Bartolomeo},
  journal={arXiv preprint arXiv:2411.15717},
  year={2024}
}

@article{cai2018simple,
  title={A simple yet effective baseline for non-attributed graph classification},
  author={Cai, Chen and Wang, Yusu},
  journal={arXiv preprint arXiv:1811.03508},
  year={2018}
}

@article{chen2025greedy,
  title={Greedy low-rank gradient compression for distributed learning with convergence guarantees},
  author={Chen, Chuyan and He, Yutong and Li, Pengrui and Jia, Weichen and Yuan, Kun},
  journal={IEEE Transactions on Signal Processing},
  year={2026},
  publisher={IEEE}
}

@software{flax2020github,
  author = {Jonathan Heek and Anselm Levskaya and Avital Oliver and Marvin Ritter and Bertrand Rondepierre and Andreas Steiner and Marc van {Z}ee},
  title = {{F}lax: A neural network library and ecosystem for {JAX}},
  url = {http://github.com/google/flax},
  version = {0.11.1},
  year = {2024},
}

@software{jraph2020github,
  author = {Jonathan Godwin and Thomas Keck and Peter Battaglia and Victor Bapst and Thomas Kipf and Yujia Li and Kimberly Stachenfeld and Petar Veli\v{c}kovi\'{c} and Alvaro Sanchez-Gonzalez},
  title = {{J}raph: {A} library for graph neural networks in jax.},
  url = {http://github.com/deepmind/jraph},
  version = {0.0.1.dev},
  year = {2020},
}

@article{monga2021algorithm,
  title={Algorithm unrolling: Interpretable, efficient deep learning for signal and image processing},
  author={Monga, Vishal and Li, Yuelong and Eldar, Yonina C},
  journal={IEEE Signal Processing Magazine},
  volume={38},
  number={2},
  pages={18--44},
  year={2021},
  publisher={IEEE}
}

@article{boyd2011,
    year = {2011},
    volume = {3},
    journal = {Foundations and Trends in Machine Learning},
    title = {Distributed Optimization and Statistical Learning via the Alternating Direction Method of Multipliers},
    issn = {1935-8237},
    number = {1},
    pages = {1-122},
    author = {Stephen Boyd and Neal Parikh and Eric Chu and Borja Peleato and Jonathan Eckstein}
}

@inproceedings{NEURIPS2021_afdec700,
 author = {Ichnowski, Jeffrey and Jain, Paras and Stellato, Bartolomeo and Banjac, Goran and Luo, Michael and Borrelli, Francesco and Gonzalez, Joseph E and Stoica, Ion and Goldberg, Ken},
 booktitle = {Advances in Neural Information Processing Systems},
 pages = {21043--21055},
 title = {Accelerating Quadratic Optimization with Reinforcement Learning},
 volume = {34},
 year = {2021}
}

@article{amos2023tutorial,
  title={Tutorial on amortized optimization},
  author={Amos, Brandon},
  journal={Foundations and Trends in Machine Learning},
  volume={16},
  number={5},
  pages={592--732},
  year={2023},
  publisher={Now Publishers, Inc.}
}

@article{chen2022learning,
  title={Learning to optimize: A primer and a benchmark},
  author={Chen, Tianlong and Chen, Xiaohan and Chen, Wuyang and Heaton, Howard and Liu, Jialin and Wang, Zhangyang and Yin, Wotao},
  journal={Journal of Machine Learning Research},
  volume={23},
  number={189},
  pages={1--59},
  year={2022}
}

@article{he2024mathematics,
  title={A Mathematics-Inspired Learning-to-Optimize Framework for Decentralized Optimization},
  author={He, Yutong and Shang, Qiulin and Huang, Xinmeng and Liu, Jialin and Yuan, Kun},
  journal={arXiv preprint arXiv:2410.01700},
  year={2024}
}

@article{battaglia2018relational,
  title={Relational inductive biases, deep learning, and graph networks},
  author={Battaglia, Peter W and Hamrick, Jessica B and Bapst, Victor and Sanchez-Gonzalez, Alvaro and Zambaldi, Vinicius and Malinowski, Mateusz and Tacchetti, Andrea and Raposo, David and Santoro, Adam and Faulkner, Ryan and others},
  journal={arXiv preprint arXiv:1806.01261},
  year={2018}
}

@inproceedings{qian2024exploring,
  title={Exploring the power of graph neural networks in solving linear optimization problems},
  author={Qian, Chendi and Ch{\'e}telat, Didier and Morris, Christopher},
  booktitle={International Conference on Artificial Intelligence and Statistics},
  pages={1432--1440},
  year={2024},
  organization={PMLR}
}

@article{sjolund2022factorization,
  title={Graph-based Neural Acceleration for Nonnegative Matrix Factorization}, 
  author={Jens Sjölund and Maria Bånkestad},
  year={2022},
  journal={arXiv preprint arXiv:2202.00264},
}

@InProceedings{zhang2014lasso,
  title = 	 {Asynchronous Distributed ADMM for Consensus Optimization},
  author = 	 {Zhang, Ruiliang and Kwok, James},
  booktitle = 	 {Proceedings of the 31st International Conference on Machine Learning},
  pages = 	 {1701--1709},
  year = 	 {2014},
  editor = 	 {Xing, Eric P. and Jebara, Tony},
  volume = 	 {32},
  number =       {2},
  series = 	 {Proceedings of Machine Learning Research},
  address = 	 {Bejing, China},
  month = 	 {22--24 Jun},
  publisher =    {PMLR}
}

@article{makh2017admm,
author = {Makhdoumi, Ali and Ozdaglar, Asuman},
issn = {0018-9286},
journal = {IEEE transactions on automatic control},
volume = {62},
year = {2017},
language = {eng},
number = {10},
pages = {5082-5095},
publisher = {IEEE},
title = {Convergence Rate of Distributed {ADMM} Over Networks},
}

@inproceedings{gilmer2017neural,
  title={Neural message passing for quantum chemistry},
  author={Gilmer, Justin and Schoenholz, Samuel S and Riley, Patrick F and Vinyals, Oriol and Dahl, George E},
  booktitle={International conference on machine learning},
  pages={1263--1272},
  year={2017},
  organization={PMLR}
}

@article{hausner2024neural,
  title={Neural incomplete factorization: learning preconditioners for the conjugate gradient method},
  author={Paul H{\"a}usner and Ozan {\"O}ktem and Jens Sj{\"o}lund},
  journal={Transactions on Machine Learning Research},
  issn={2835-8856},
  year={2024},
}

@article{teixeira2015admm,
  title={The ADMM algorithm for distributed quadratic problems: Parameter selection and constraint preconditioning},
  author={Teixeira, Andre and Ghadimi, Euhanna and Shames, Iman and Sandberg, Henrik and Johansson, Mikael},
  journal={IEEE Transactions on Signal Processing},
  volume={64},
  number={2},
  pages={290--305},
  year={2015},
  publisher={IEEE}
}

@article{fahy2024greedy,
  title={Greedy Learning to Optimize with Convergence Guarantees},
  author={Fahy, Patrick and Golbabaee, Mohammad and Ehrhardt, Matthias J},
  journal={arXiv preprint arXiv:2406.00260},
  year={2024}
}

@article{andrychowicz2016learning,
  title={Learning to learn by gradient descent by gradient descent},
  author={Andrychowicz, Marcin and Denil, Misha and Gomez, Sergio and Hoffman, Matthew W and Pfau, David and Schaul, Tom and Shillingford, Brendan and De Freitas, Nando},
  journal={Advances in neural information processing systems},
  volume={29},
  year={2016}
}

@article{scarselli2008graph,
  title={The graph neural network model},
  author={Scarselli, Franco and Gori, Marco and Tsoi, Ah Chung and Hagenbuchner, Markus and Monfardini, Gabriele},
  journal={IEEE transactions on neural networks},
  volume={20},
  number={1},
  pages={61--80},
  year={2008},
  publisher={IEEE}
}

@article{gao2025gradient,
  title={Gradient Methods with Online Scaling Part I. Theoretical Foundations},
  author={Gao, Wenzhi and Chu, Ya-Chi and Ye, Yinyu and Udell, Madeleine},
  journal={arXiv preprint arXiv:2505.23081},
  year={2025}
}

@article{Gabay,
author = {Gabay, Daniel and Mercier, Bertrand},
copyright = {1976},
issn = {0898-1221},
journal = {Computers \& mathematics with applications (1987)},
volume = {2},
year = {1976},
language = {eng},
number = {1},
pages = {17-40},
publisher = {Elsevier Ltd},
title = {A dual algorithm for the solution of nonlinear variational problems via finite element approximation},
}

@article{Glowinski,
author = {Glowinski, R. and Marroco, A.},
issn = {0397-9342},
journal = {Revue française d'automatique, informatique, recherche opérationnelle. Analyse numérique},
volume = {9},
year = {1975},
language = {fre ; eng},
number = {R2},
pages = {41-76},
publisher = {EDP Sciences},
title = {Sur l'approximation, par éléments finis d'ordre un, et la résolution, par pénalisation-dualité d'une classe de problèmes de Dirichlet non linéaires},
}

@article{Erdos:1959:pmd,
  author = {Erd\"os, P and R\'enyi, A},
  journal = {Publicationes Mathematicae Debrecen},
  pages = {290--297},
  title = {On Random Graphs I},
  volume = 6,
  year = 1959
}

@article{yang2020distributed,
  title={Distributed least squares solver for network linear equations},
  author={Yang, Tao and George, Jemin and Qin, Jiahu and Yi, Xinlei and Wu, Junfeng},
  journal={Automatica},
  volume={113},
  pages={108798},
  year={2020},
  publisher={Elsevier}
}

@article{velivckovic2021neural,
  title={Neural algorithmic reasoning},
  author={Veli{\v{c}}kovi{\'c}, Petar and Blundell, Charles},
  journal={Patterns},
  volume={2},
  number={7},
  year={2021},
  publisher={Elsevier}
}

@inproceedings{amos2017optnet,
  title={Optnet: Differentiable optimization as a layer in neural networks},
  author={Amos, Brandon and Kolter, J Zico},
  booktitle={International conference on machine learning},
  pages={136--145},
  year={2017},
  organization={PMLR}
}

@article{sambharya2024learning,
  title={Learning to warm-start fixed-point optimization algorithms},
  author={Sambharya, Rajiv and Hall, Georgina and Amos, Brandon and Stellato, Bartolomeo},
  journal={Journal of Machine Learning Research},
  volume={25},
  number={166},
  pages={1--46},
  year={2024}
}

@InProceedings{hausner2025learning,
  title={Learning incomplete factorization preconditioners for {GMRES}},
  author={H{\"a}usner, Paul and Nieto Juscafresa, Aleix and Sj{\"o}lund, Jens},
  booktitle={Proceedings of the 6th Northern Lights Deep Learning Conference (NLDL)},
  pages={85--99},
  year={2025},
  volume={265},
  series={Proceedings of Machine Learning Research},
  publisher={PMLR},
}

@InProceedings{chen2024expressive,
  title = 	 {Expressive Power of Graph Neural Networks for ({M}ixed-Integer) Quadratic Programs},
  author =       {Chen, Ziang and Chen, Xiaohan and Liu, Jialin and Wang, Xinshang and Yin, Wotao},
  booktitle = 	 {Proceedings of the 42nd International Conference on Machine Learning},
  year = 	 {2025},
  volume = 	 {267},
  series = 	 {Proceedings of Machine Learning Research},
  month = 	 {13--19 Jul},
  publisher =    {PMLR},
}

@inproceedings{saravanos2024deep,
    title={Deep Distributed Optimization for Large-Scale Quadratic Programming},
    author={Augustinos D Saravanos and Hunter Kuperman and Alex Oshin and Arshiya Taj Abdul and Vincent Pacelli and Evangelos Theodorou},
    booktitle={The Thirteenth International Conference on Learning Representations},
    year={2025},
}

@inproceedings{arisaka2023principled,
  title={Principled acceleration of iterative numerical methods using machine learning},
  author={Arisaka, Sohei and Li, Qianxiao},
  booktitle={International Conference on Machine Learning},
  pages={1041--1059},
  year={2023},
  organization={PMLR}
}

@article{noah2024distributed,
  title={Distributed learn-to-optimize: Limited communications optimization over networks via deep unfolded distributed {ADMM}},
  author={Noah, Yoav and Shlezinger, Nir},
  journal={IEEE Transactions on Mobile Computing},
  year={2024},
  publisher={IEEE}
}

@article{barber2024node-level_alpha,
  author  = {Rina Foygel Barber and Emil Y. Sidky},
  title   = {Convergence for nonconvex ADMM, with applications to CT imaging},
  journal = {Journal of Machine Learning Research},
  year    = {2024},
  volume  = {25},
  number  = {38},
  pages   = {1--46},
}

@incollection{eckstein1990admm,
    author ={Jonathan Eckstein and Dimitri P. Bertsekas},
    title = {An Alternating Direction Method for Linear Programming},
    publisher = {Laboratory for Information and Decision Systems, Massachusetts Institute of Technology},
    year = {1990}
}

@article{banert2024accelerated,
  title={Accelerated forward-backward optimization using deep learning},
  author={Banert, Sebastian and Rudzusika, Jevgenija and {\"O}ktem, Ozan and Adler, Jonas},
  journal={SIAM Journal on Optimization},
  volume={34},
  number={2},
  pages={1236--1263},
  year={2024},
  publisher={SIAM}
}

@article{zhang2019accelerating,
  title={Accelerating ADMM for efficient simulation and optimization},
  author={Zhang, Juyong and Peng, Yue and Ouyang, Wenqing and Deng, Bailin},
  journal={ACM Transactions on Graphics (TOG)},
  volume={38},
  number={6},
  pages={1--21},
  year={2019},
  publisher={ACM New York, NY, USA}
}

@article{ulyanov2016instance,
  title={Instance normalization: The missing ingredient for fast stylization},
  author={Ulyanov, Dmitry and Vedaldi, Andrea and Lempitsky, Victor},
  journal={arXiv preprint arXiv:1607.08022},
  year={2016}
}

@article{he2000alternating,
  title={Alternating direction method with self-adaptive penalty parameters for monotone variational inequalities},
  author={He, Bing-Sheng and Yang, Hai and Wang, SL},
  journal={Journal of Optimization Theory and applications},
  volume={106},
  number={2},
  pages={337--356},
  year={2000},
  publisher={Springer}
}

@inproceedings{li2023learning,
  title={Learning to optimize distributed optimization: Admm-based dc-opf case study},
  author={Li, Meiyi and Kolouri, Soheil and Mohammadi, Javad},
  booktitle={2023 IEEE Power \& Energy Society General Meeting (PESGM)},
  pages={1--5},
  year={2023},
  organization={IEEE}
}

\newpage
\appendix

\section{Glossary of Notation}
\label{app:notation}
\begin{table}[ht]
  \centering
  \caption{Summary of the notation used throughout the paper.}
  \label{tab:notation}
  \resizebox{\textwidth}{!}{
  \begin{tabular}{cll}
    \toprule
    Symbol & Description & Comments \\
    \midrule
    $\mathcal{G} = (\mathcal{V}, \mathcal{E})$ & Graph with set of vertices $\mathcal{V}$ and edges $\mathcal{E}$ & $\mathcal{V} = \{1, 2, \dots, m \}$ and $\mathcal{E} \subseteq \mathcal{V} \times \mathcal{V}$ \\
    $\mathcal{N}(i)$ & Set of neighboring nodes of $i$ & By default we assume $i \notin \mathcal{N}(i)$ \\
    $d_i$ & Degree of node $i$ & Equal to $| \mathcal{N}(i) |$\\
    $D, \tilde{D}$ & Degree matrix and weighted degree matrix \\
    $m$ & Number of agents / nodes in the graph \\
    $n$ & Dimension of the local optimization problem \\
    $f_i$ & Local objective function of node / agent $i$ \\
    $b_i, B_i$ & Local objective function information \\
    $x_i$ & Local optimization variable of node / agent $i$ \\
    \midrule
    $ I_n$ & Identity matrix of size $n$ \\
    $1_n$ & Vector of all ones of size $n$ \\
    \midrule
    ${V} = [v_i]_{i=1}^m$ & Node features in GNN \\
    ${E} = [e_{ij}]$ for $(i, j) \in \mathcal{E}$ & Edge features in GNN & Typically implemented as a sparse matrix \\
    $l$ & Layer in GNN & Twice as many GNN layers as ADMM steps \\
    $\theta$ & All learnable network parameters \\
    $\mathcal{D}$ & Dataset used for training \\
    $\varepsilon$ & Small constant to avoid division by zero \\
    \midrule
    $\alpha$ & ADMM step size &  \\
    $\mathcal{L}$ & Weighted Laplacian matrix & $\mathcal{L} = D - E$ \\
    $P$ & Communication matrix \\
    $x, z$ & ADMM primal variables & Stacked and slack distributed variables \\
    $y$ & ADMM distributed variable & This is a reformulation of the $z$ variable \\
    $\lambda$ & ADMM dual variable \\
    $r, s$ & Primal and dual ADMM residuals \\
    $\mu, \tau$ & Adaptive step size parameters \\
    $A, B$ & ADMM linear constraint matrices \\
    $k$ & Iterations of the ADMM algorithm \\
    $K$ & Total number of ADMM steps in the unrolling \\
    $L$ & Number of steps to switch to fixed step size & Often $L = K$\\
    \bottomrule
  \end{tabular}
  }
\end{table}

\section{Problem details}
\label{app:problems}
Here, we describe in more detail how we generate the different problem classes used in the experiments in Section~\ref{sec:experiments} and describe how to generate ground-truth labels for them which are used for the supervised training as well as provide details on how the subproblem in equation~\eqref{eq:xproblem} is solved in practice.
Further, we provide the additional information to describe how the distribution of problems is generated for each problem.

Across all problem domains, the underlying graph structure is based on the Erdős-Rényi random graph model \citep{Erdos:1959:pmd} with $m=8$ nodes.
The edge probability is fixed to $p=0.4$ in the experiments.
We discard sampled graphs that are not connected.
Moreover, the dimensionality of the solution $x^\star\in\mathbb R^n$ is fixed at $n=2$ in all experiments.

For each problem class, we generate a total of $900$ instances for training, $100$ instances for validation and $100$ instances for testing purposes.
All results in the main text are evaluated on the testing instances which are not seen by the network during training.

\subsection{Network average consensus}
For the network average consensus problem, each node or agent $i\in \{1,\ldots,m\}$ in the network has a local objective function that minimizes the distance between the local optimization variable $x_i$ and its local private data $b_i$.

\paragraph{Problem generation}
This local objective function is parameterized for every agent by a vector $b_i \in \mathbb{R}^n$. The vector is sampled independently from a $n$-dimensional normal distribution $\mathcal N(0,\Sigma)$ with covariance matrix $\Sigma=100 I_n$.

\paragraph{Subproblem solution}
The solution to the subproblem \eqref{eq:xproblem} for this use-case is a special case of the result presented for the distributed least-squares problem below with $B_i = I_n$.
Since the resulting linear system for the problem that we derive in equation~\eqref{eq:closedform} is diagonal, the problem can be solved more efficiently in practice requiring only $\mathcal{O}(n)$ steps and does not require expensive matrix inversion techniques.
Instead, we can take the reciprocal of the diagonal elements to obtain the inverse.
Note that these elements are always non-zero by the construction of the linear system, and thus the inverse is well-defined.

\paragraph{Global solution}
For training, we assume access to the global solution $x^\star$ for all problem instances.
We can efficiently generate these solutions by utilizing the underlying problem structure and the direct access to all problem data.
The minimizer is given by
\begin{equation}
    x^\star = \frac1n \sum_{i=1}^n b_i
\end{equation}
In other words, the agents are collaborating to find the mean of their vectors $b_i$ without sharing the actual data.

\subsection{Distributed least squares problem}
In the distributed least squares problem, the local objective function solves the least-squares problem for the privately available data. In addition to the data vector $b_i$ from the network average consensus problem, each agent also has a private feature transformation matrix $B_i\in\mathbb R^{n\times n}$.
The latter is used to weight the linear regression coefficients of $x_i$ before comparing with $b_i$.

\paragraph{Problem generation}
The local objective function is parametrized for every agent by the pair $(B_i,b_i)\in \mathbb R^{n\times n}\times \mathbb R^n$. Each entry of the matrix $B_i$ is sampled from a continuous uniform distribution $\mathcal U[0,1]$ on the unit interval. In addition, we assume in our experiments that $B_i$ is of full rank to obtain better convergence behavior of ADMM and ensure closed-form solutions. We enforce this by discarding and resampling $B_i$ in case the magnitude of an eigenvalue is below $1/10$. The data vector $b_i$ is sampled as before.

\paragraph{Subproblem solution}
In the distributed least squares problem, the solution to the optimization problem to update the $x$-variable from \eqref{eq:xproblem} has a closed form solution and can be solved as the solution to the following linear equation system:
\begin{equation}
\label{eq:closedform}
x_i^{k+1}=(2B_i^TB_i+\alpha M_{ii})^{-1}\bigl(2B_i^Tb_i-P_{ii}\lambda_i^k-\bar \lambda_{\rightarrow i}^k+\alpha (M_{ii}x_i^k-P_{ii}y_i^k-\bar y^k_{\rightarrow i})\bigl),
\end{equation} where $M_{ii}$ is a diagonal matrix defined by
\begin{equation}
\begin{aligned}
    M_{ii}:&=\left(\sum_{j\in \mathcal N(i)}P_{ji}^TP_{ji}+P_{ii}^TP_{ii}\right)\otimes  I_n\\
    &=\Biggl(\sum_{j\in\mathcal N(i)}e_{ji}^2+\bigl(\underbrace{\sum_{j\in\mathcal N(i)}e_{ji}}_{\text{weighted degree of $i$}}\bigl)^2\Biggl)\otimes  \,  I_n
\end{aligned}
\end{equation}
since $P_{ii}=\tilde D_{ii}=\sum_{j\in \mathcal{N}(i)}e_{ji}$ is the weighted degree in the case where we choose the weighted Laplacian as our communication matrix for the algorithm.
Note that the (weighted) degree and the sum of the weighted connections can be easily computed as a message in the MPNN.

We can observe in particular that the matrix here is positive definite since $B_i^TB_i$ is positive (semi)-definite and $M_{ii}$ is a diagonal matrix with positive diagonal entries.
Therefore, we can use the iterative conjugate gradient method to compute the solution to the equation system which scales computationally better than inverting or factorizing the matrix for large-scale problems.

\paragraph{Global solution}
The global minimizer is given by
\begin{equation}
    x^\star=\left(\sum_{i=1}^mB_i^TB_i\right)^{-1}\left(\sum_{i=1}^mB_i^Tb_i\right)\,.
\end{equation}
Note that the inverse exists since again $B_i^TB_i$ is positive definite for every $i$, due to the construction of the training data, and hence is the sum described in the problem generation paragraph.

\section{Implementation details}
\label{app:training}
In the following, we discuss the details of the training scheme and the chosen hyperparameters and tuning steps in order to ensure reproducibility of our presented results.

\paragraph{Framework}
We use the JAX framework \citep{jax2018github} to implement the differentiable distributed ADMM algorithm as a message-passing network.
In particular, we are using the \texttt{jraph} toolkit to implement the message-passing steps \citep{jraph2020github} and integrate it with the \texttt{flax} library to design neural network components using the novel \texttt{linen} API \citep{flax2020github}.

\paragraph{Optimization subproblem}
There is one obvious difficulty in making the GNN fully differentiable.
In each update of the node features $x_i$, an optimization problem of the form \eqref{eq:node_update1} needs to be solved, and the gradient of the network parameters $\theta$ needs to be computed with respect to the solution to this optimization problem.
In the case of step size learning, all of $\alpha_i$, the previous iterates, and incoming messages depend on the parameters of the network through previous updates.
To still compute the gradient, we need to differentiate the solution to the optimization problem with respect to the parameters of the network $\frac{\partial x_i^{k+1}}{\partial \theta}$ which can be decomposed via the chain rule $\frac{\partial x_i^{k+1}}{\partial \alpha_i} \cdot \frac{\partial \alpha_i}{\partial \theta}$.

Computing the first part can be achieved in closed form by the means of implicit differentiation~\citep{amos2017optnet}.
However, implicit differentiation can in practice be costly and requires additional overhead.
Instead, we solve the subproblem \eqref{eq:node_update1} approximately by unrolling an iterative solver\footnote{Here, we unroll the iterative solver within each unrolled iteration of ADMM.} or use the closed-form solution to \eqref{eq:node_update1} if it exists. In both cases, we compute its gradient using standard automatic differentiation techniques.
Along the same lines, the partial derivative of the optimization problem~\eqref{eq:xproblem} with respect to the learned edge weights $e_{ij}$ is computed.
The remaining update steps can be differentiated using standard automatic differentiation tools provided by JAX \citep{jax2018github}.

\paragraph{Network architectures}
For the network components to predict the hyperparameters, within the message-passing steps, we are using a 2-layer multi-perceptrons.
The hidden layer has $32$ neurons and there is $1$ output which gets passed through a $\texttt{softplus}$ activation function to ensure positivity.
We use a \texttt{ReLU} non-linearity between the two layers.

\begin{itemize}
    \item We use a single network for all iterations from $k=2$ until $K=10$. For the first step, we do not use learning but use the default step size. Thus, there are a total of $9$ learned steps in the unrolling scheme. The input to the network is given by the stacked optimization variables of the current iteration. Thus, the total number of trainable parameters is $3\thinspace753$.
    \item The input for the edge weight prediction is given by the concatenated local degree profiles of the adjacent nodes \citep{cai2018simple}. Each of these consists of $5$ features containing statistics about the the neighborhood structure including the degree, maximum and minimum neighboring degree, as well as the mean and variance. To ensure that the communication matrix is symmetric, we sum up the edge predictions. Further, we use the same predicted communication matrix for all iterations. Thus, the number of parameters in the model is $385$.
    \item In the combined model, we use both the step size network and the edge weight network together. Thus, the total number of parameters is $4\thinspace138$.
\end{itemize}

\paragraph{Training hyperparameters}
For the training, we are using the Adam optimizer with a learning rate of $10^{-4}$ and global gradient clipping of with radius $1.0$.
We are training the model for $100$ epochs using a batch size of $5$ via gradient accumulation.
Thus, during training a total of $18\thinspace000$ parameter update steps are executed.
To avoid division by zero, we include the hyperparameter $\varepsilon$ into the loss function in \eqref{eq:loss}.
In the experiments, we set  $\varepsilon=10^{-5}$.
\paragraph{ADMM hyperparameters} If not learned or further specified, we apply the naive step size $\alpha=1$ and the Laplacian matrix $\mathcal L$ of the graph as the communication matrix $P$ in all algorithm steps. Like the Laplacian, the degree $d_i$ of each node $i$ depends on the respective graph instance.

\paragraph{ADMM initialization}
In our experiments, we initialize all $x_i$ by $\mathbf 0 \in \mathbb R^n$ and hence also all $y_i$ by $\mathbf 0 \in \mathbb R^n$. While it is generally possible to use warm starting with the proposed framework, e.g. through learning \citep{sambharya2024learning}, additional algorithm steps are required in this case to ensure that the consensus variables $y_i$ are appropriately initialized. For details, see lines 2-6 in Algorithm~\ref{alg:basic_admm}.

\paragraph{Fixed baseline}
The fixed baseline method uses a constant, global $\alpha$ across all iterations, which is chosen via grid search on the validation data.
We allow $\alpha$ to be in the interval $[0.001, 10]$ using linear spacing for $100$ different values.\footnote{We choose the initial interval such that it covers the empirically observed learned step sizes across the unrolled iterations.}
This is a form of black-box optimization. 
To choose the $\alpha$ that performs best on average for all instances, we use the same objective as in training, defined in equation~\eqref{eq:loss}. For both problem classes, we select the $\alpha$ in our grid that performs best in this metric on the validation data after $K=10$ iterations, and apply it on the test data in the reported results.
This leads to selecting $\alpha=1.112$ for the consensus problem, and $\alpha = 1.011$ for the least squares problem.
This approach is similar to the method proposed by \citet{noah2024distributed} in the case where only one constant step size is learned, but it relies on black-box optimization rather than gradient-based optimization to determine the parameters.

\paragraph{Adaptive baseline}
In the adaptive baseline, the local node-level step sizes $\alpha_i$ gets adjusted dynamically based on the balance of primal and dual residuals \citep{boyd2011}.
Intuitively, if the primal residual is too large, we increase $\alpha$ to shift the focus towards enforcing the consensus, while if the dual residual is too large, we decrease the step size to allow for more flexibility in the local node updates.
We utilize the common heuristic introduced by \citet{he2000alternating} to update the step size in each step as:
\begin{equation}
    \alpha^{k+1} = \begin{cases}
        \tau \alpha^k \quad & \text{if } \| r^k \| > \mu \|s^k\|, \\
        \tau^{-1} \alpha^k \quad&  \text{if } \|s^k\| > \mu \|r^k\|, \\
        \alpha^k \quad & \text{otherwise}.
    \end{cases}
\end{equation}
Here, $\tau \in \mathbb{R}_+$ and $\mu \in \mathbb{R}_+$ are parameters of the adaptive scheme.
Typical default choices are $\tau = 2$ and $\mu = 10$ \citep{he2000alternating}.
The primal and dual residual at iteration $k$ are denoted by $r^k$ and $s^k$ respectively and defined as:
\begin{equation}
    \begin{aligned}
        r_{ij}^{k+1} &= P_{ij} x_j^{k+1} - z_{ij}^{k+1} = y_i^{k+1}, \\
        s_j^{k+1} &= \alpha \sum_{i=1}^m \left[
        P_{ij}^2 \left(x_j^{k+1} - x_j^k\right) +
         P_{ij} \left(y_i^{k+1} - y_i^k\right)\right] \\
         & = \alpha \left(\left[ \sum_{i=1}^m P_{ij}^2 \right] (x_j^{k+1} - x_j^k) + d_j (y^{k+1}_j - y^{k}_j) + \bar{y}_{\rightarrow j}^{\,k+1} - \bar{y}_{\rightarrow j}^{\,k}  \right)
    \end{aligned}
\end{equation}
The second formulation of the dual residual is readily available in our message-passing framework.
Note that the primal residual is an edge-based version of the newly introduced $y^k$ in the distributed ADMM framework as defined in equation~\eqref{eq:ylamproblem}, measuring the constraint violation of the primal problem. The dual residual is a measure of change in this auxiliary variable and the local solution $x_i$. In order to compute a node-based primal residual for predicting a node-based $\alpha_i$ with the step size heuristic, we use
\begin{equation}
    \| r_i \|_2 = \sqrt{\sum_{j\in \mathcal N(i)}r_{ij}^Tr_{ij}}=\sqrt{d_i \,y_i^Ty_i}=\sqrt{d_i}\,\|y_i\|_2.
\end{equation}

As for the fixed baseline, we choose the hyperparameters of the adaptive scheme using grid search over the validation data, based on average performance after $K=10$ iterations. 
For the $\mu$ parameter, that controls how frequently the step size parameter is changed, we search over the choices $\{1, 5, 10, 15, 20, 25, 30, 25, 30, 35, 40\}$.
For the scaling parameter, that adjusts the step size we pick values in the interval $(1, 2]$.
The grid search picks $20$ different values from the interval with logarithmic spacing.
Thus, in total we evaluate $200$ different combinations of parameters for the tuning of the adaptive step size scheme.
From this, we obtain the optimal values $\mu=10$ and $\tau=1.32$ for the consensus problem, and similarly $\mu = 10$ and $\tau = 1.36$ for the least-squares problem.
In order to avoid numerical breakdowns, we reset the step size to the fixed value of $\alpha=1$ after the first $K=10$ iterations, as in the other methods.

\paragraph{Computational resources}
All reported experiments are executed on a single NVIDIA TITAN Xp GPU with 12~GB of memory.
Training a single epoch requires $10$ minutes plus additional time for the validation process.
In total, the training of one model for $100$ epochs requires around $20$ hours.

\section{Additional results}
\label{app:results}
Here, we provide additional results to complement the empirical evaluation of our methods based on the convergence time, the distribution of solving times, and the out-of-domain (OOD) performances.

\begin{table}[b!]
    \centering
    \caption{Timing and convergence comparison of the different methods until the relative objective reaches a threshold $\varepsilon$. The total computational budget for each problem is $1$ second. For each solver we check how many instances achieve the desired accuracy within this budget and report the resulting success rate ($\uparrow$ higher is better). For the successfully solved instances per method, we measure the average number of iterations and required time in seconds ($\downarrow$ lower is better).
    Best is \textbf{bold}, second best \underline{underlined}.}
    \label{tab:times}
    \resizebox{\textwidth}{!}{
        \begin{tabular}{lccccccccc}
        \toprule
        & \multicolumn{3}{c}{$\varepsilon = 0.05$} & \multicolumn{3}{c}{$\varepsilon = 0.01$} & \multicolumn{3}{c}{$\varepsilon = 0.001$} \\
        \cmidrule(lr){2-4} \cmidrule(lr){5-7} \cmidrule(lr){8-10}
        \textbf{Consensus} & Iterations & Time & Success & Iterations & Time & Success & Iterations & Time & Success \\
        \midrule
        Fixed & 11.64 & 0.23 & 0.94 & 16.36 & 0.30 & 0.87 & 29.42 & 0.50 & 0.76 \\
        Adaptive & 11.30 & 0.30 & 0.93 & 15.76 & 0.36 & 0.83 & 26.85 & 0.57 & 0.71 \\
        Global $\alpha$ & 7.82 & 0.20 & \underline{0.96} & 19.37 & 0.39 & 0.91 & 35.31 & 0.64 & 0.78 \\
        Local $\alpha_i$ & \textbf{5.61} & \textbf{0.16} & \underline{0.96} & \underline{12.72} & \underline{0.28} & 0.89 & \underline{26.88} & 0.51 & 0.80 \\
        Weight $e_{ij}$ & 11.19 & 0.24 & \underline{0.96} & 15.62 & 0.32 & \underline{0.92} & 27.38 & \underline{0.46} & \underline{0.82} \\
        Combined & \textbf{5.61} & \underline{0.17} & \textbf{0.98} & \textbf{10.47} & \textbf{0.25} & \textbf{0.96} & \textbf{23.56} & \textbf{0.44} & \textbf{0.90} \\
        \midrule
        \textbf{Least-squares} & Iterations & Time & Success & Iterations & Time & Success & Iterations & Time & Success \\
        \midrule
        Fixed & 9.61 & 0.30 & 0.85 & 14.72 & 0.44 & 0.60 & 21.23 & 0.62 & 0.26 \\
        Adaptive & 9.20 & 0.32 & 0.85 & 13.29 & 0.44 & 0.59 & 20.23 & 0.62 & 0.26 \\
        Global $\alpha$ & \textbf{5.31} & \underline{0.22} & \textbf{0.94} & \underline{11.30} & \underline{0.36} & \underline{0.76} & \underline{18.47} & 0.58 & 0.34 \\
        Local $\alpha_i$ & 6.27 & \underline{0.22} & 0.89 & 11.67 & 0.39 & \underline{0.76} & 18.97 & 0.60 & 0.35 \\
        Weight $e_{ij}$ & 10.88 & 0.33 & 0.86 & 15.68 & 0.47 & 0.63 & 19.54 & \underline{0.56} & \underline{0.39} \\
        Combined & \underline{5.41} & \textbf{0.21} & \textbf{0.94} & \textbf{10.13} & \textbf{0.34} & \textbf{0.83} & \textbf{17.90} & \textbf{0.53} & \textbf{0.41} \\
        \bottomrule
    \end{tabular}
    }
\end{table}

\subsection{ADMM convergence and solver time}
In Table~\ref{tab:times}, we compare the different methods based on achieving a certain threshold in relative objective value within a given computational budget of one second. For every instance on which a certain method reaches the threshold within the budget, we measure the required number of iterations and wall-clock time. In the table, we report these values as averages across successful instances per method. Further, we report the fraction of successfully solved instances within the budget.
Note that this leads to a dependent metric since the iterations and time are conditional on solving. Thus, we need to evaluate the different methods based on their Pareto front instead of looking at the metrics individually.
A method is on the Pareto front if no other method achieves both a higher success rate and lower iteration/time performance at the same time.

Across both problem classes, all learned solvers are able to solve more instances within the computational budget than the fixed and adaptive baseline. Moreover, for every threshold, there exist multiple learned methods that are Pareto improvements to the two baselines, i.e. more instances are solved while requiring fewer average iterations and solving them in faster average time. In particular, the combined method appears to be the strongest choice as it is always a Pareto improvement over the baselines, solves the most instances, and is on the Pareto front for all thresholds and problems. Only for $\varepsilon=0.05$ there exists a learned method that requires marginally fewer iterations or less time.

\begin{figure}[b!]
    \centering
    \includegraphics[width=\linewidth]{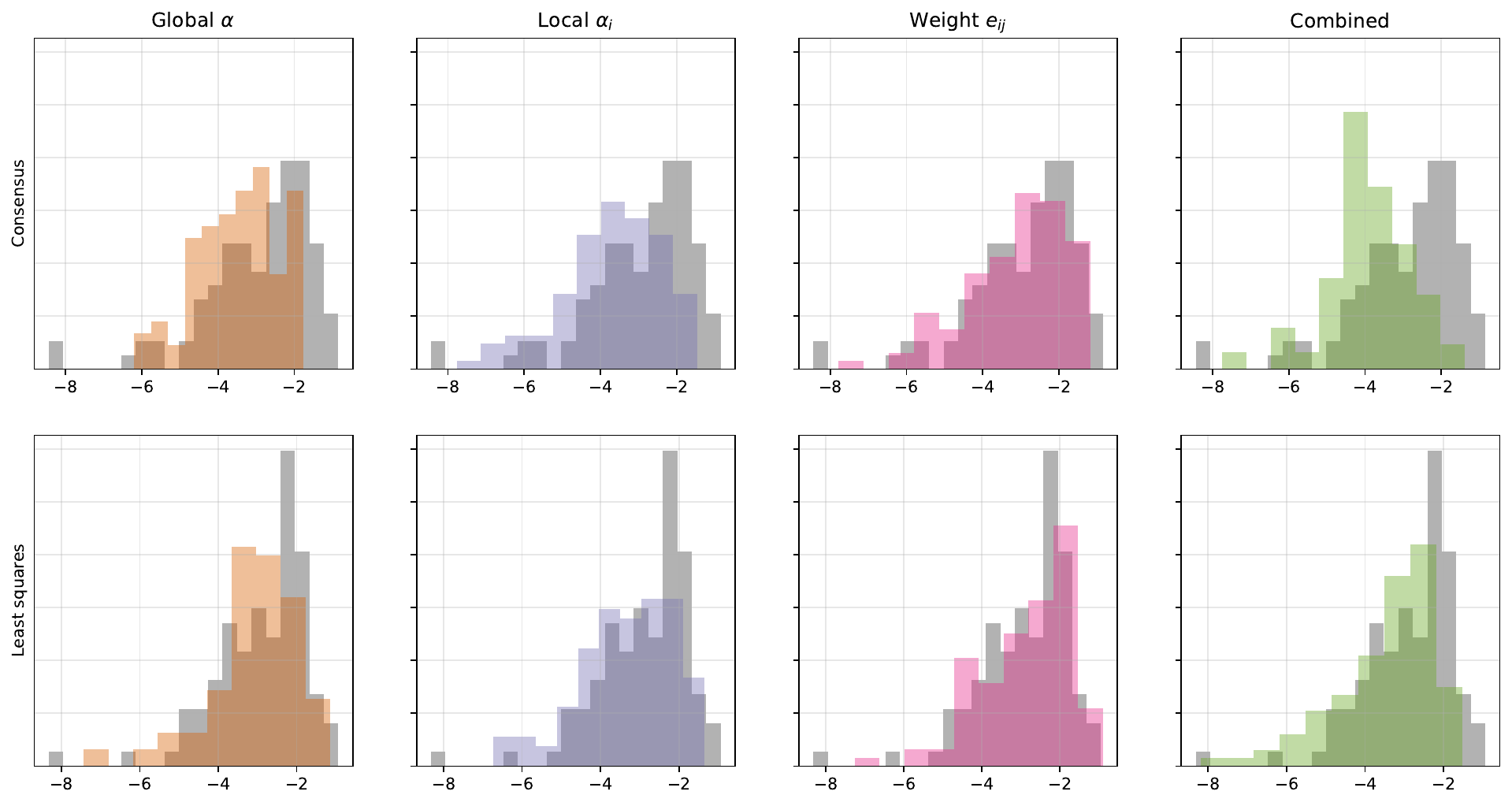}
    \caption{Pairwise comparison of the the distributions of normalized objective value across the test set between the fixed baseline method (gray) and the learned methods (colored). The x-axis shows the log-scaled normalized residual after $10$ iterations.}
    \label{fig:distr}
\end{figure}

\subsection{Distribution of solving times}
In the previous results, we only report the mean performance across all problem instances to summarize the performance in a single number.
Here, we additionally show the distribution of the performance across the $100$ individual test instances.
For each learned ADMM solver, we compare the distribution of relative objectives after $K=10$ iterations in Figure~\ref{fig:distr}.
Overall, we can see that there is a large spread of relative objectives achieved after the iterations showing that the problems are vastly different in difficulty.

One observation is that the learned ADMM method can lead to slightly worse performance for individual instances for which the fixed baseline method works very well.
This is due to the fact that the training aims to improve the mean performance across all instances \citep{arisaka2023principled}.

\subsection{Generalization behavior of learned methods}
\paragraph{OOD generalization}
To examine the robustness and generalization ability of our learned methods, we evaluate their performance on larger graphs unseen during training, and compare the results with the behavior of the two baselines. For the out-of-distribution samples, we slightly change the data distribution by adjusting the edge weight parameter $p$ in the random graph generation to keep the approximate neighborhood structure fixed.
Given the initial value for $p=0.4$ the expected number of neighbors in a graph with $m=8$ nodes is $2.8$.
We adjust the edge probability for a graph with $m \in \{ 16, 32, 64, 128 \}$ nodes to $p = 2.8 / (m-1)$.
Thus, the expected degree of each node remains the same even when increasing the graph size.
This is more realistic than scaling the degree of the nodes as well, as in practical applications, the degree often represents a physical communication limit, which does not increase when more neighbours are in the network.

The obtained results are shown in Table~\ref{tab:ood}. Although the network is only trained on small graphs with $m=8$ nodes, we observe that all learned methods generalize well and outperform the baselines even on the OOD data. The performance of all methods drops compared to the original data shown in Table~\ref{tab:resultsconsensus}, but does not continue to worsen when increasing the graph size further.

\begin{table}[t!]
    \centering
    \caption{
    Results for OOD generalization with increasing number of nodes in the distributed system after $K=10$ steps. For each size, we solve $20$ problem instances and report the mean performance. All models are trained using problem instances with $m=8$ nodes. Best method is \textbf{bold}, second best \underline{underlined}.}
    \label{tab:ood}
    \resizebox{\textwidth}{!}{
    \begin{tabular}{lcccccccccc}
    \toprule
    & \multicolumn{2}{c}{m=16} & \multicolumn{2}{c}{m=32} & \multicolumn{2}{c}{m=64} & \multicolumn{2}{c}{m=128} \\
    \cmidrule(lr){2-3} \cmidrule(lr){4-5} \cmidrule(lr){6-7} \cmidrule(lr){8-9}
    \textbf{Consensus} & Error & Consensus & Error & Consensus & Error & Consensus & Error & Consensus \\
    \midrule
    Fixed & 22.92 & 21.26 & 18.04 & 17.24 & 16.90 & 16.51 & 20.07 & 19.88 \\
    Adaptive & 25.01 & 24.00 & 21.37 & 20.89 & 20.56 & 20.36 & 24.58 & 24.48 \\
    Global $\alpha$ & 18.91 & 18.16 & 16.11 & 15.82 & 16.30 & 16.15 & 19.19 & 19.13 \\
    Local $\alpha_i$ & \underline{15.90} & \underline{15.74} & \underline{13.16} & \underline{13.04} & \underline{13.74} & \underline{13.69} & \underline{16.37} & \underline{16.34} \\
    Weight $e_{ij}$ & 19.63 & 17.90 & 17.99 & 17.29 & 16.21 & 15.88 & 19.76 & 19.59 \\
    Combined & \textbf{12.93} & \textbf{12.47} & \textbf{10.91} & \textbf{10.58} & \textbf{11.65} & \textbf{11.54} & \textbf{13.95} & \textbf{13.90} \\
    \bottomrule
    \end{tabular}
    }
\end{table}

\begin{figure}[h]
    \centering
    \includegraphics[width=\linewidth]{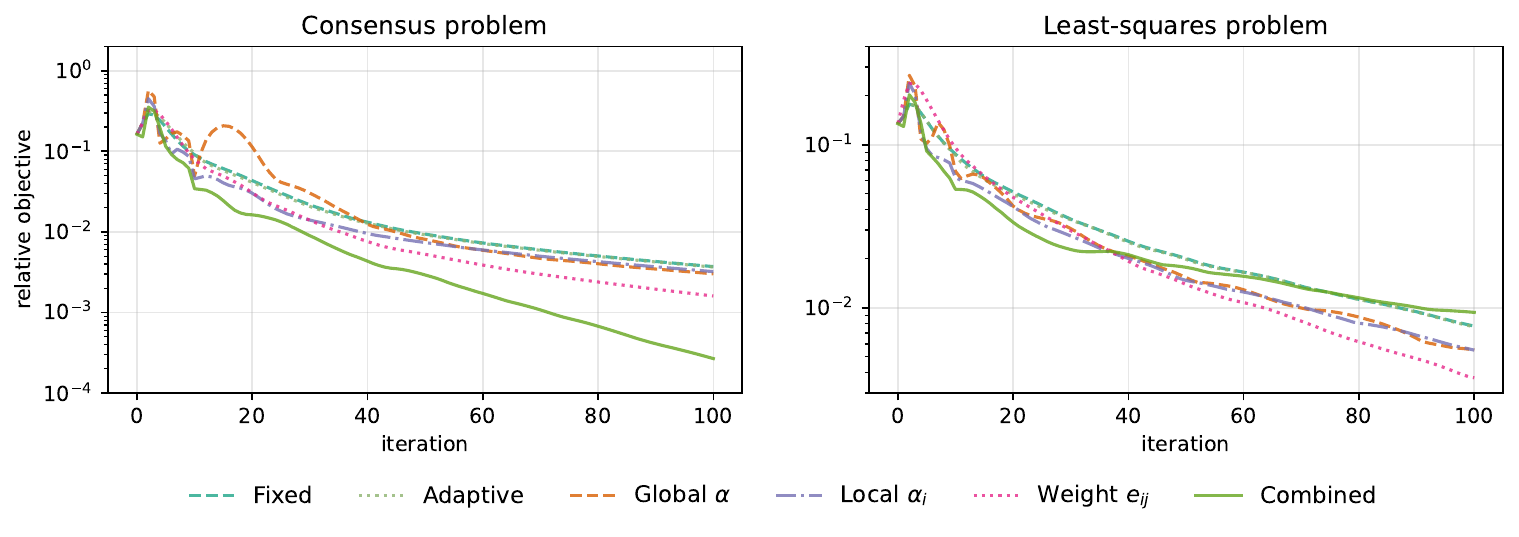}
    \caption{Semi-log plots of the mean relative objective over all $100$ test instances in the two problem classes for $100$ iterations. Note that this is far beyond the trained unrolling of $K=10$ iterations.}
    \label{fig:iterationsood}
\end{figure}

\paragraph{Additional ADMM iterations}
In order to show the performance of our method when increasing the number of iterations even further beyond the training regime, we showcase the performance for $100$ iterations in Figure~\ref{fig:iterationsood}. Figure~\ref{fig:function} showed the same results until $20$ iterations are reached.

The results show that the learned methods continue to outperform the baselines, even far beyond the trained iterations. The only exception to this is the combined method in the least-squares problem, which slows down after 50 iterations. Particularly noteworthy is the strong performance of the method with learned communication matrix, as the learned matrix is used in all iterations and not only in the trained ones.
However, using the learned communication matrix together with the learned step sizes performs best in the consensus example while achieving the worst in the least squares problem.

\newpage

\section{Algorithms}\label{sec:algorithms}
\label{app:alg}
\subsection{Decentralized, distributed ADMM}In the following, we state the decentralized, distributed ADMM algorithm (Algorithm \ref{alg:basic_admm}) from \citet{makh2017admm} adapted to our notation. As described in Section \ref{sec:admm}, each iteration updates all three iterates $(x_i,y_i,\lambda_i)$ and includes two communication steps to pass information about the current iterates to the neighbors in $\mathcal G$.\par

\begin{algorithm}[h!]
\caption{Decentralized, distributed ADMM}\label{alg:basic_admm}
\begin{algorithmic}[1]
\State \textbf{Hyperparameters:} $\alpha>0$, communication matrix $P$ ($=\mathcal L$ here), $d_i$ degree of node $i$
\State \textbf{Initialize} $x_i^0 \in \mathbb{R}^n, \lambda_i^0=0\in \mathbb{R}^n$ for $i=1,\ldots,m$, and $k=0$
\State Communicate $x_i^0$ to all neighbors $j\in \mathcal N(i)$
\For{$i=1,\ldots,m$}
    \State Initialize $y_i^0$
\begin{equation*}
        y_i^0=\frac{1}{d_i+1}\sum_{j\in \mathcal N(i)\cup\{i\}}P_{ij}x_j^0
    \end{equation*}
\EndFor
\While{stopping criterion is not reached}
\For{$i = 1,\ldots,m$ \textbf{in parallel}}
    \State Communicate $y_i^{k}$ and $\lambda_i^k$ to all neighbors $j\in \mathcal N(i)$
    \State Update local solution candidate $x$:
    \begin{align*}
            x_i^{k+1}=\underset{x_i}{\text{arg min}}\,\,\left(f_i(x_i)+\sum_{j\in \mathcal N(i)\cup\{i\}}\left(( \lambda_{j}^{k})^T(P_{ji}x_i) +\frac{\alpha}{2}\|P_{ji}(x_i-x_i^k)+y_{j}^k\|_2^2\right)\right)\,.
        \end{align*}
    \State Communicate $x_i^{k+1}$ to all neighbors $j\in \mathcal N(i)$
    \State Update local deviation-variable $y$:
        \begin{equation*}
            y_i^{k+1}=\frac{1}{d_i+1}\sum_{j\in \mathcal N(i)\cup\{i\}}P_{ij}x_j^{k+1}\,.
        \end{equation*}
    \State Update local dual-variable $\lambda$:
        \begin{equation*}
            \lambda_i^{k+1}=\lambda_i^k+\alpha y_i^{k+1}\,.
        \end{equation*}
           
    \EndFor
    \State $k\gets k+1$
\EndWhile
\State \textbf{Return:} $x^{k}=[\,x_i^k\,]_{i=1}^m$
\end{algorithmic}
\end{algorithm}

\subsection{Message-passing formulation of distributed ADMM} When it comes to defining a message-passing network that performs one iteration of Algorithm \ref{alg:basic_admm}, it is critical to reformulate the update steps in such a way that they only require knowledge of the aggregation of incoming messages and not of all individual messages. We pointed this out in Section \ref{sec:gnn-admm} and defined respective \texttt{message}, \texttt{aggregate}, and \texttt{update} functions that respect the MPNN framework. The two resulting message-passing steps are presented in Algorithm \ref{alg:dec_ADMM_gnn}. To highlight the one-to-one correspondence, we compare it with the equivalent Algorithm \ref{alg:dec_ADMM}, which is identical to a single iteration of the original distributed ADMM algorithm (see lines of 8-14 of Algorithm \ref{alg:basic_admm}).

\paragraph{Equivalence between updates.} For completeness, we will briefly show the equivalence between the original $x$-update step from \eqref{eq:xproblem} and the rewritten step in \eqref{eq:node_update1} here. Details for the $y$ and $\lambda$ update are omitted since the equivalence follows immediately from the definition of the messages.

Equation \eqref{eq:xproblem} (or line 8 of Algorithm \ref{alg:basic_admm}) includes a sum over features from neighboring nodes that we need to revise. The first summand in this sum can be rewritten easily by the definition of the aggregated messages $\bar{\lambda}_{\rightarrow i}^{\,k}$:
\begin{align}
    \sum_{j\in \mathcal N(i)\cup\{i\}}(\lambda_j^k)^T(P_{ji}x_i)=
    (P_{ii}\lambda_i^k)^Tx_i+\Biggl(\underbrace{\sum_{j\in \mathcal N(i)}P_{ji}^T\lambda_j^k}_{=:\,\bar{\lambda}_{\rightarrow i}^{\,k}}\Biggl)^Tx_i\,,
\end{align}
Note that $P_{ji}$ must be included in the message since weighting is no longer possible once all $\lambda_j^k$ are aggregated. The second summand, the penalty term for consensus, splits into the following
\begin{align}
    \sum_{j\in \mathcal N(i)\cup\{i\}}\|P_{ji}(x_i-x_i^k)+y_j^k\|_2^2=\sum_{j\in \mathcal N(i)\cup\{i\}}\Biggl(\|&P_{ji}(x_i-
    x_i^k)\|_2^2\notag\\&\underbrace{+\|y_j^k\|_2^2+2(P_{ji}(x_i-x_i^k))^Ty_j^k}_{\text{includes features from neighbors}}\,\Biggl)\,.
\end{align}
Since the $x$-update optimizes the equation with respect to $x_i$, we understand that the quadratic term $\|y_j^k\|_2^2$ is irrelevant for this update. The mixed term, which involves features from neighboring nodes, splits as follows
\begin{align}
    2 \cdot \sum_{j\in \mathcal N(i)\cup\{i\}}(P_{ji}(x_i-x_i^k))^Ty_j^k=2(x_i-x_i^k)^T\Biggl(P_{ii}^Ty_i^k+\underbrace{\sum_{j\in \mathcal N(i)}P_{ji}^Ty_j^k}_{=: \,\bar y_{\rightarrow i}^{\,k}}\Biggl)\,,
\end{align}
where $\bar y_{\rightarrow i}^{\,k}\in \mathbb R^n$ is the desired message containing the $y$-features of all neighbors from node $i$. By the same argument as above, we can leave the constant term involving $x_i^k$ out in the optimization. 

Note that this rule can be modified further if the edge weights $e_{ij}$ are not fixed over the whole algorithm and are not accessible for node $i$. In this case, one must add $(P_{ji},P_{ji}^2)$ to the message.

\paragraph{MLP Integration.}
For step size learning (both global and node-level), we can predict and use a new step size in each iteration. To achieve this, we insert each MLP that predicts a new step size into the first node \texttt{update} function of each iteration, before the $x$ variable is updated. In Algorithm \ref{alg:dec_ADMM_gnn}, this is exactly between lines 5 and 6 of the pseudocode. 
The predicted step size is then used during an entire iteration.

Conversely, in the edge-level task, we predict the edge-weights $e_{ij}$ only once and keep them fixed during the whole $K=10$ unrolled iterations. The corresponding MLP is placed before the first iteration of Algorithm \ref{alg:dec_ADMM_gnn}.
\newpage

\scalebox{0.81}{\centering
\begin{minipage}{\linewidth}
\centering
    \begin{algorithm}[H]
\renewcommand{\thealgorithm}{2.1}
\caption{One iteration of decentralized, distributed ADMM}\label{alg:dec_ADMM}
\begin{algorithmic}[1]
\For{$i \in \mathcal V$ \textbf{in parallel}}
\State
\begin{tcolorbox}[cleanframebox=white,colback=red!3, borderline={0.25mm}{-0.0mm}{red}]
\State Communicate $y_i^{k}$ and $\lambda_i^k$ to all neighbors $j\in \mathcal N(i)$
    \State Update local solution candidate $x$:
    \begin{align*}
            x_i^{k+1}=\underset{x_i}{\text{arg min}}\,\biggl(&f_i(x_i)\,\\&+\hspace{-0.1cm}\underset{j\in \mathcal N(i)\cup\{i\}}{\sum}\hspace{-0.1cm}\left(( \lambda_{j}^{k})^T(P_{ji}x_i) +\frac{\alpha}{2}\|P_{ji}(x_i-x_i^k)+y_{j}^k\|_2^2\right)\biggl)\,.
        \end{align*}
\end{tcolorbox}
\vspace{0.2cm}
    \begin{tcolorbox}[cleanframebox=white,borderline={0.25mm}{-0.0mm}{dashed,blue},colback=blue!3]
    \State Communicate $x_i^{k+1}$ to all neighbors $j\in \mathcal N(i)$
    \State Update local deviation-variable $y$:
        \begin{equation*}
            y_i^{k+1}=\frac{1}{d_i+1}\sum_{j\in \mathcal N(i)\cup\{i\}}P_{ij}x_j^{k+1}\,.
        \end{equation*}
    \State Update local dual-variable $\lambda$:
        \begin{equation*}
            \lambda_i^{k+1}=\lambda_i^k+\alpha y_i^{k+1}\,.
        \end{equation*}
           
     \end{tcolorbox}
    \EndFor
\end{algorithmic}
\end{algorithm}
    \begin{algorithm}[H]
\renewcommand{\thealgorithm}{2.2}\caption{One iteration of decentralized, distributed ADMM \textbf{as 2-block GNN}}\label{alg:dec_ADMM_gnn}
\begin{algorithmic}[1]
     \begin{tcolorbox}[cleanframebox=white,colback=red!3, left=4pt,right=4pt,top=4pt,bottom=4pt, borderline={0.25mm}{-0.0mm}{red}]\For{$(j,i)\in \mathcal E$ \textbf{in parallel}}
        \State $m_{ji} = \left(P_{ji}\lambda_j^k,P_{ji}y_j^k\right)$
        \Comment{\texttt{message} 1\,}
    \EndFor
    \For {$i\in \mathcal V$ \textbf{in parallel}} 
        \State $\left(\bar\lambda_{\rightarrow i}^k,\bar y_{\rightarrow i}^k\right)  = \sum_{(j,i)\in \mathcal E}m_{ji}$
        \Comment{\texttt{aggregation} 1\,}
        \State \hspace{0.67cm}${v}_i \hspace{0.79cm} = \left(x_i^{k+1},y_i^k,\lambda_i^k\right),\,\text{where } $
        \Comment{\texttt{update} 1\,}
        
        \begin{align*}\label{eq:node_update1_app}
     x_i^{k+1}=\underset{x_i}{\text{arg min}}\,\,\biggl(f_i(x_i)+\bigl(P_{ii}\lambda_i^k+&\bar \lambda_{\rightarrow i}^{\,k} +\,\alpha(P_{ii}y_i^k+\bar y_{\rightarrow i}^{\,k}) \bigl)^Tx_i \notag\\
     &+\frac{\alpha}{2}\sum_{j\in \mathcal N(i)\cup\{i\}}\|P_{ji}(x_i-x_i^k)\|_2^2\biggr)\,.
        \end{align*}
        \EndFor \end{tcolorbox}\vspace{0.2cm}
    \begin{tcolorbox}[cleanframebox=white,borderline={0.25mm}{-0.0mm}{dashed,blue},colback=blue!3,left=4pt,right=4pt,top=4pt,bottom=4pt]
    \For {$(j,i)\in \mathcal E$ \textbf{in parallel}}
        \State ${m}_{ji} = P_{ij}x_j^{k+1}$
        \Comment{\texttt{message} 2\,}
    \EndFor
    \For {$i\in \mathcal V$ \textbf{in parallel}}
         \State $\bar x_{\rightarrow i}^{k+1} = \sum_{(j,i)\in \mathcal E}m_{ji}$
        \Comment{\texttt{aggregation} 2\,}
        \State \hspace{0.2cm}${v}_i \hspace{0.28cm} = \left(x_i^{k+1},y_i^{k+1},\lambda_i^{k+1}\right),\,\text{where } $ \Comment{\texttt{update} 2\,}
        \begin{equation*}\label{eq:node_update2}
    \begin{aligned}
        y_i^{k+1}&=\frac {1}{d_i+1}\left(\bar x_{\rightarrow i}^{k+1}+P_{ii}x_i^{k+1}\right)\,,\\
        \lambda_i^{k+1}&=\lambda_i^k+\alpha y_i^{k+1}\,.
    \end{aligned}
    \end{equation*}
       
    \EndFor
    \end{tcolorbox}
\end{algorithmic}
\end{algorithm}
    \end{minipage}
}

\end{document}